\documentclass[a4paper,fleqn]{cas-dc}
\usepackage[numbers, sort&compress]{natbib}
\def\tsc#1{\csdef{#1}{\textsc{\lowercase{#1}}\xspace}}
\tsc{WGM}
\tsc{QE}

\usepackage{url}

\usepackage[colorlinks]{hyperref} 
\usepackage{amsmath,amssymb,amsfonts}
\usepackage{algorithmic}
\usepackage{graphicx}
\usepackage{textcomp}
\usepackage{bm,multirow, booktabs}
\usepackage[ruled]{algorithm2e}

\usepackage{marginnote}

\begin{document}
\let\WriteBookmarks\relax
\def\floatpagepagefraction{1}
\def\textpagefraction{.001}

\shorttitle{JotlasNet}    

\shortauthors{Y. Zhang \textit{et~al.}}  

\title[mode=title]{JotlasNet: Joint Tensor Low-Rank and Attention-based Sparse Unrolling Network for Accelerating Dynamic MRI}%

\author[1]{Yinghao Zhang}[orcid=0000-0001-8501-6180]
\author[2]{Haiyan Gui}
\author[2]{Ningdi Yang}
\author[1]{Yue Hu}[orcid=0000-0002-4648-611X]
\cormark[1]
\cortext[cor1]{Corresponding author: e-mail: huyue@hit.edu.cn}
\address[1]{School of Electronics and Information Engineering, Harbin Institute of Technology, Harbin, China}
\address[2]{The Fourth Hospital of Harbin, Harbin, China}

\begin{abstract}
	Joint low-rank and sparse unrolling networks have shown superior performance in dynamic MRI reconstruction. 
	% However, existing unrolling networks are mainly based on single priors, while the works on composite priors are relatively limited and face complex network structures. \
	However, existing works mainly utilized matrix low-rank priors, neglecting the tensor characteristics of dynamic MRI images, and only a global threshold is applied for the sparse constraint to the multi-channel data, limiting the flexibility of the network. Additionally, most of them have inherently complex network structure, with intricate interactions among variables. 
	% However, existing unrolling networks lack sufficient utilization of high-dimensional low-rank information and thorough exploitation of sparse priors. The structure of joint low-rank and sparse networks is inherently complex, with intricate interactions among variables. 
	In this paper, we propose a novel deep unrolling network, JotlasNet, for dynamic MRI reconstruction by jointly utilizing tensor low-rank and attention-based sparse priors. 
	Specifically, we utilize tensor low-rank prior to exploit the structural correlations in high-dimensional data. Convolutional neural networks are used to adaptively learn the low-rank and sparse transform domains. 	
	A novel attention-based soft thresholding operator is proposed to assign a unique learnable threshold to each channel of the data in the CNN-learned sparse domain.
	The network is unrolled from the elaborately designed composite splitting algorithm and thus features a simple yet efficient parallel structure.  Extensive experiments on two datasets (OCMR, CMRxRecon) demonstrate the superior performance of JotlasNet in dynamic MRI reconstruction. 
\end{abstract}

% keywords can be removed

\maketitle

\section{Introduction}
% 加速磁共振成像已经成为社区内非常重要的研究方向。对于能够展现时间维度变化信息的动态磁共振成像，需要采集的数据量更大，其对于减少成像时间的需求更加迫切，而且所需要的加速倍数也更高。现如今，以低于奈奎斯特采样的方式采集k空间数据是加速MRI的主要手段，但将不可避免地导致图像混叠伪影, 加速和图像质量之间存在trade-off。设计能够在固定加速倍数下提高图像质量，或者换句话说，能够在图像质量满足临床需要的前提下容忍更高的加速倍数的重建算法至关重要。
Accelerating magnetic resonance imaging (MRI) has become a highly crucial research direction in the MRI research community \cite{muckley2021results}. For dynamic MRI that captures temporal changes, a larger amount of data needs to be collected. The demand for reducing imaging time is more urgent in this context, and the required acceleration factor is also higher. Currently, acquiring k-space data below the Nyquist sampling rate is the primary approach for accelerating MRI \cite{lustig2008compressed,lustig2007sparse}. However, this undersampling strategy inevitably leads to image aliasing artifacts, creating a trade-off between acceleration and image quality. Therefore, it is paramount to design reconstruction algorithms that can enhance image quality at a fixed acceleration factor or, in other words, tolerate higher acceleration factors while meeting clinical requirements for image quality.

% 基于迭代优化算法的MRI重建方法在近十几年取得了较好的效果。具体来说，图像的先验信息作为约束项与k空间数据保真项合并构建优化重建模型，然后，ADMM、ISTA等迭代求解算法被应用来求解这一优化问题。稀疏先验作为压缩感知的核心思想之一，被广泛应用于MRI重建中。在各种变换域内的稀疏先验，如小波、全变分、时域FFT和字典表示等，都在加速MRI应用中取得了很好的效果。另外，动态磁共振图像作为真实物理空间的时间序列图像，在时间维度上变化往往较为缓慢。时间上的相关性较高，信息冗余较大。卡索拉低展开矩阵的低秩先验也被采用来提取和利用时间帧的相关性来重建图像。随着研究的深入，三维张量低秩模型也被提出用于动态MRI重建，并且因为其能够额外利用图像空间的相关性和对高维数据结构相关性的精确建模而优越于卡索拉低展开矩阵低秩模型。进一步地，低秩和稀疏先验结合的方法也陆续被提出，并且证明了能够取得比单一先验的重建方法更为优越的效果。这些方法主要分为两类，一类是低秩+稀疏的方案，即将动态磁共振图像分为运动缓慢的低秩背景成分和运动快速的稀疏成分；另一类是低秩和稀疏的联合模型，即将低秩和稀疏约束同时施加到待重建目标图像上以充分挖掘图像本身的数据先验特征，此类方法也是我们本文所关注的重点。上述优化方法虽然在近十几年取得了卓越的成绩，然而他们普遍存在以下几个问题：1）收敛所需迭代步数一般较多，导致重建时间较长；2）算法的超参数往往根据小数据集的重建效果手动调整，缺乏鲁棒性且耗费时间。
In recent decades, iterative optimization algorithms have emerged as promising tools for accelerating MRI \cite{lustig2008compressed}. Specifically, explicit prior about MRI images is incorporated as a constraint term, combined with the fidelity term of k-space data, to construct an optimization reconstruction model. Iterative algorithms like ADMM (Alternating Direction Method of Multipliers) \cite{boyd2011distributed,afonso2010admm} and ISTA (Iterative Shrinkage-Thresholding Algorithm) \cite{beck2009fast} are commonly employed to tackle these optimization problems. Sparse priors, a fundamental aspect of compressive sensing, have been widely employed in MRI reconstruction \cite{lustig2007sparse,huang2011efficient,jung2009k,ravishankar2010mr}. Additionally, methods leveraging matrix/tensor low-rank priors \cite{liang2007spatiotemporal,he2016accelerated} have been proposed. Moreover, hybrid approaches combining low-rank and sparse priors have demonstrated superior performance compared to single-prior methods. These hybrids typically fall into two categories: one involves a low-rank plus sparse scheme \cite{otazo2015low}, where dynamic MRI images are partitioned into slowly varying low-rank backgrounds and rapidly changing sparse components; the other is a joint low-rank and sparse formulation \cite{lingala2011accelerated}, simultaneously applying low-rank and sparse constraints to the target image to fully exploit data prior features. We focus our discussion on the latter category in this paper.
Despite the notable progress of these optimization methods, they face several common challenges: \textbf{a.} convergence often requires a considerable number of iterations, resulting in lengthy reconstruction times; \textbf{b.} manual tuning of hyperparameters based on small dataset reconstructions is common, lacking robustness and time-consuming.

% Unrolling网络能够有效地克服上述问题，其将固定步数的迭代优化算法展开成深度学习网络结构，使得原迭代算法中的超参数变为网络中的可学习参数，进而能够通过端到端的监督学习方式在大数据集上进行训练，避免了人工调参的问题。同时，与深度学习网络相同，在经过充分训练后，Unrolling网络能够在测试阶段能够以极快的速度（秒级）进行重建。另外，由于其能够利用数据显式先验以及基于迭代优化算法展开的特征，Unrolling网络还具有很强的可解释性，这是传统深度学习网络所不具备的。在MRI重建领域，Unrolling网络已经被证明现如今表现最为优异的重建方法。
Deep unrolling networks (DUNs) \cite{LeCun2010unroll} effectively address the aforementioned issues by unfolding fixed-number iterations of the optimization algorithm into a deep learning network. This transformation turns the hyperparameters of the original iterative algorithm into learnable parameters within the network. Subsequently, these parameters can be trained in an end-to-end supervised manner on large datasets, eliminating the need for manual tuning. Furthermore, DUNs can leverage neural networks such as CNN (convolutional neural network) to learn a suitable transform domain, enhancing reconstruction accuracy by more efficiently utilizing the explicit priors. After sufficient training, DUNs can achieve rapid reconstruction at test time (within seconds) \cite{ref_slrnet,ref_DCCNN}. Additionally, due to their ability to leverage explicit data priors and their unfolded feature, DUNs exhibit strong interpretability—an attribute not typically found in traditional deep learning networks. To date, DUNs have proven to be the most outstanding reconstruction method in the field of MRI reconstruction \cite{muckley2021results}. However, current DUNs for accelerating dynamic MRI face some challenges.
\textbf{a.} Tensor low-rank priors have demonstrated the ability to efficiently exploit high-dimensional structural similarities in optimization-based methods, thereby achieving superior performance. However, in joint low-rank and sparse DUNs, the low-rank priors are typically derived from the Casorati matrix, neglecting the tensor characteristics of dynamic MRI images.
% \textbf{a.} Current networks fall short in effectively utilizing low-rank priors. They inherit the optimization-based methods of extracting temporal correlation in dynamic MRI images using matrix low-rank, while neglecting spatial low-rank characteristics and the exploitation of the high-dimensional structure of dynamic MRI tensors; 
\textbf{b.} The two CNNs (see LR or S layer in Fig.\ref{fig:network}) that learn the transform domain for efficiently utilizing explicit priors, are usually constrained to be inverse, limiting the flexibility of the network.
\textbf{c.} In the CNN-learned sparse transform domain, data is represented as multi-channel features. However, current networks \cite{ref_ISTANET,ref_slrnet} typically apply the same thresholding operation to all channels using traditional soft thresholding (ST) \cite{donoho1995noising} operators.
\textbf{d.} Current iterative optimization algorithms for joint low-rank and sparse unrolling networks \cite{ref_slrnet} are highly complex and involve numerous temporary variables and intricate interactions. This complexity renders the design of composite-prior DUNs difficult and cumbersome, while also hindering efficient backpropagation to some extent, consequently leading to suboptimal reconstruction results.

% 用于加速MRI的深度展开网络大抵可以通过其利用的数据先验类型进行粗略界定。利用CNN学习的隐式图像先验的展开网络取得了不错的效果，然而这种隐式约束导致优化模型无法明确构建，从而使得网络可解释性变差。低秩和稀疏作为MRI重建中的重要显示先验，与其结合的展开网络也被提出。这些网络大多都利用CNN等神经网络根据数据自适应地学习一个稀疏或低秩变换域，然后在该变换域内进行相应的先验约束。如此，摆脱了人工设计变换域的困难同时也能够根据数据更好的提取相应的先验特征，进而能够取得更好地表现。另外，类似传统迭代优化方法，低秩和稀疏结合的方法同样被提出并且报道了比单一先验更好的重建效果。例如，低秩plus稀疏的LplusS-Net和joint低秩和稀疏的SLR-Net。然而，目前的展开网络存在一些问题：1）对于CNN学习的稀疏先验的利用不足，主要表现在，在CNN学习的变换域内，数据表现为多通道的特征，然而，目前的网络往往通过传统的软阈值算子对所有通道进行相同的阈值操作。2）对于低秩和稀疏的联合模型，网络结构复杂使得训练困难，从而导致较差的重建效果。这主要是因为，用于网络展开的迭代优化算法较为复杂，引入的临时变量众多且交互错综复杂，阻碍了高效的反向传播。
% 相比较于传统的张量CP和TUCKER分解，在t-SVD范式下的具有张量核范数的严格定义，其能够在理论上严格证明是张量秩的凸包络。t-SVD的计算可以通过沿某个维度的多个矩阵的SVD来实现，不需要像CP分解一样的迭代优化过程，进而显著地减小了计算量。最近的变换域t-SVD的提出，使得CNN-learned变换域低秩先验具备了理论支撑，进而能够利用CNN从数据中提取高效的低秩信息。
The tensor nuclear norm \cite{lu2019tensor} based on tensor singular value decomposition (t-SVD) \cite{ref_tsvd} effectively leverages the low-rank characteristics of tensors and has demonstrated excellent performance across various applications \cite{zhang2024t2lr,zhang2020video}.
Compared to traditional tensor CP and Tucker decompositions \cite{kolda2009tensor}, t-SVD offers a rigorously defined tensor nuclear norm, which can be theoretically proven to be the convex envelope of the tensor rank. The computation of t-SVD can be achieved by performing SVD on multiple matrices along a specific dimension, eliminating the need for the iterative optimization process required by CP decomposition, thereby significantly reducing computational complexity.
The recent development of transformd t-SVD \cite{zhang2024t2lr} has provided theoretical support for the low-rank prior learned in the CNN transform domain. This advancement allows CNNs to effectively extract low-rank information from data, leveraging the low-rank prior in a theoretically sound manner.
Additionally, the composite splitting algorithm \cite{huang2011composite} for models with joint dual priors, directly utilizes the ISTA algorithm for solving, resulting in a simple algorithmic structure. Moreover, it has been demonstrated to achieve faster convergence speed and better reconstruction performance compared to complex ADMM-based algorithms \cite{huang2011composite}.

Inspired by these findings, in this paper, we proposed a novel JotlasNet: joint tensor low-rank and attention-based sparse unrolling network for accelerating dynamic MRI, with the following contributions for the aforementioned challenges.
% 1. 提出利用张量低秩和张量稀疏先验联合进行构建动态MRI重建深度展开网络，其中，低秩和稀疏变换域由CNN来自适应地学习，张量低秩先验基于张量奇异值分解框架下的张量核范数构建以利用高维数据的结构相关性，张量稀疏先验基于张量L1范数构建。2. 提出一种基于注意力的软阈值算子（AST），为CNN稀疏变换域中的多通道数据的每个通道单独分配一个可学习的阈值参数，增强了稀疏约束的自由度，以提高重建质量。3. 设计一种基于Composite Splitting algorithm的求解joint Low-rank and Sparse重建模型的迭代求解算法。该算法具有同单一先验模型类似的简单且高效的结构，进而使得由其展开的JotlasNet能够得到更好的重建效果。4. 在两个公开数据集上（OCMR, CMRxRecon）进行充足的实验，证明了JotlasNet网络在动态MRI重建中的优越性能。
\begin{itemize}
	\item Propose a novel network incorporating tensor low-rank and attention-based sparse priors for dynamic MRI reconstruction. The tensor low-rank prior is derived from t-SVD, effectively exploiting the structural correlations within high-dimensional data. The CNNs adaptively learn the low-rank and sparse transforms without constraints, allowing the JotlasNet to simultaneously exploit explicit priors and implicit CNN-extracted features.
	\item Introduce an Attention-based Soft Thresholding operator (AST), assigning a learnable threshold to each channel of the data in the CNN-learned sparse domain. This enhances the flexibility of the tensor sparse constraint, thereby improving the reconstruction quality.
	\item Develop a simple yet efficient structure for the joint low-rank and sparse unrolling network based on the composite splitting algorithm. 
	\item Extensive experiments on two publicly available datasets (OCMR, CMRxRecon) demonstrate the superior performance of the proposed JotlasNet in dynamic MRI reconstruction.
\end{itemize}

% In Section \ref{sec:related}, we review related works, while Section \ref provides a detailed introduction to our proposed JotlasNet. Experimental results and discussion are presented in Sections 4 and 5, respectively. Finally, Section 6 concludes the paper.

Although our preliminary work was presented at ICIP 2023 \cite{zhang2023dusnet}, this paper represents a significant extension. Unlike our prior work, which solely relied on sparse priors, this paper introduces a network combining both low-rank and sparse priors. Furthermore, it explores the high-dimensional structural correlations in the data using tensor low-rank prior. Additionally, we have carefully designed the network structure for joint priors and enriched both experimental and theoretical aspects.

\section{Related Works}
\label{sec:related}
% % 在详细展开具体方法的论述之前，我们首先对本文中的符号进行定义，具体如表1所示。
% Before delving into a detailed exposition of the specific methods, we first establish the definitions of symbols used in this paper, as outlined in Tab.\ref{tab:symbol}.
% % Tab.generated by Excel2LaTeX from sheet 'Sheet1'
% \begin{table}[htbp]
% 	\centering
% 	\caption{The definitions of symbols used in this paper.}
% 	  \begin{tabular}{lll}
% 	  \toprule
% 	  Symbol & Format & Example \\
% 	  \midrule
% 	  Operator & sans serif capital letter & $\mathsf{A}$ \\
% 	  Number set & blackboard bold capital letter & $\mathbb{C}$ \\
% 	  Tensor & Euler script uppercase letter & $\mathcal{X}$\\
% 	  Matrix & bold capital letter & $\mathbf{X}$ \\
% 	  Vector & bold lowercase letter & $\mathbf{b}$ \\
% 	  Scalar & upper/lowercase letter & $H,n_x$ \\
% 	  \bottomrule
% 	  \end{tabular}%
% 	\label{tab:symbol}%
%   \end{table}%
  
\subsection{General Reconstruction Model of Dynamic MRI}
The acquisition of dynamic MRI can be formulated as,
\begin{equation}
	\label{eq:acq}
	\mathbf{b} = \mathsf{A} (\mathcal{X}) + \mathbf{n},
\end{equation} 
where $\mathbf{b} \in \mathbb{C}^{M}$ is the acquired k-space data, $\mathcal{X} \in \mathbb{C}^{H \times W \times T}$ is the distortion-free dynamic MRI image with $H$, $W$, and $T$ representing the spatial dimensions and temporal frames, respectively, $\mathsf{A}: \mathbb{C}^{H \times W \times T} \rightarrow \mathbb{C}^{M}$ is the acquisition operator, and $\mathbf{n} \in \mathbb{C}^{M}$ is the noise. From the above equation, it can be observed that reconstructing the clean image $\mathcal{X}$ from the acquired undersampled k-space data $\mathbf{b}$ is a typical ill-posed linear inverse problem. Specifically, based on the physical principles of MRI, operator $\mathsf{A}$ can be expressed in a more detailed manner,
\begin{equation}
	\mathsf{A} = \mathsf{F_u} \circ \mathsf{S} = \mathsf{M} \circ \mathsf{F} \circ \mathsf{S},
\end{equation}
where $\mathsf{F_u}$ is the undersampling operator, $\mathsf{S}: \mathbb{C}^{H \times W \times T} \rightarrow \mathbb{C}^{C \times H \times W \times T}$ denotes the coil sensitive maps for multi-coil MRI with $C$ coils, and the symbol $\circ$ denotes the composite operation. When the sampling points are in the Cartesian grid, $\mathsf{F_u} = \mathsf{M} \circ \mathsf{F}$ holds with $\mathsf{F}$ being the unitary Fourier transform, and $\mathsf{M}$ denoting the sampling mask. As for single-coil cases, $\mathsf{S}$ can be omitted.

Based on \eqref{eq:acq}, the reconstruction model of dynamic MRI can be formulated as,
\begin{equation}
	\label{eq:rec}
	\hat{\mathcal{X}} = \arg \min_{\mathcal{X}} \frac{1}{2} \left\| \mathsf{A} (\mathcal{X}) - \mathbf{b} \right\|_2^2 + \lambda \mathsf{\Phi}(\mathcal{X}),
\end{equation}
where the first term with $l_2$ norm is the data fidelity term, the second term with $\mathsf{\Phi}: \mathbb{C}^{H \times W \times T} \rightarrow \mathbb{R}_+$ is the regularization term that encodes the prior information of MRI images, and $\lambda$ is the balance parameter.

\subsection{Unrolling Networks using Single Prior}
% 展开网络构建一般包括三个步骤，先验驱动的重建模型，迭代求解算法，展开。实际上，前两个因素即是基于迭代优化算法的重建方法的完备构建过程，展开网络在此基础上对算法在深度学习框架中展开，并赋予其基于数据集学习并端到端训练的能力。展开网络在迭代优化算法的上层包装了一层监督机制，使得以前需要人工经验性调整的超参数（例如，(1)中的lambda）变为可学习参数，另外，还可以利用一些神经网络（FC, CNN, Transformer）等进一步增强网络的特征提取能力，进而能够取得更好的重建效果。
An unrolling network \cite{LeCun2010unroll, liang2020deep, monga2021algorithm} generally involves three factors: a prior-driven reconstruction model, an iterative solving algorithm, and unfolding. Actually, the first two factors can precisely constitute a complete optimization-based reconstruction method. 

%The unrolling network builds upon this foundation by unfolding the algorithm within a deep learning framework and endowing it with the capability for learning based on datasets and end-to-end training. The unrolling network introduces the supervision mechanism on top of the optimization-based method, transforming previously manually adjusted hyperparameters, e.g., $\lambda$ in \eqref{eq:rec}, into trainable parameters. Additionally, it leverages neural networks such as Fully Connected (FC), Convolutional Neural Network (CNN), Transformer, etc., to further enhance the network's feature extraction capabilities, thereby achieving superior reconstruction results.

% 近年来，各种基于单一先验的展开网络层出不穷。
In recent years, various unrolling reconstruction networks based on a single prior have emerged and become mainstream \cite{muckley2021results}. DCCNN \cite{ref_DCCNN}, MoDL \cite{aggarwal2018modl}, and E2EVarNet \cite{sriram2020end} are representative examples that using CNNs to learn the implicit image prior. HUMUS-Net \cite{fabian2022humus} took the vision Transformer to substitute CNN. 
CineVN \cite{vornehm2025cinevn} employed a spatiotemporal E2EVarNet combined with conjugate gradient descent for optimized data consistency and improved image quality for dynamic MRI.
However, the implicit nature leads to a lack of interpretability. ISTA-Net \cite{ref_ISTANET} and FISTA-Net \cite{xiang2021fista} exploited the learned sparse prior and adopted the ST operator \cite{donoho1995noising} to enforce the sparsity constraint in a CNN-learned multi-channel sparse domain. However, the ST operator applied the same threshold to all channels and the two CNNs that learn the sparse transforms are constrained to be inverse, limiting the flexibility of the sparse constraint. 
Moreover, T2LR-Net \cite{zhang2024t2lr,zhang2023tlr} utilized the tensor low-rank prior and the tensor nuclear norm \cite{ref_tsvd} to exploit the structural correlations in high-dimensional data, in which CNNs were also used to adaptively learn the low-rank transform domain. % The above is from the view of the intrinsic prior behind DUNs. 

As for the underlying iterative solving algorithms that we termed as the structure of the DUNs, algorithms like ADMM \cite{boyd2011distributed}, ISTA \cite{beck2009fast}, and others are widely used. However, although ADMM is an effective algorithm for a lot of optimization problems, it usually requires the conjugate gradient (CG) method to solve the subproblems, especially for the MRI reconstruction cases with non-Cartesian sampling and multi-coil data. Inserting CG in DUNs will increase the complexity of the network and make the training process more difficult \cite{aggarwal2018modl}. ISTA does not require CG iterations; it can be iteratively solved through a gradient descent step and a projection step. Another widely used approach is to directly simplify the gradient descent step in ISTA through the use of the Data Consistency (DC) layer \cite{ref_DCCNN,sriram2020end}. % During the iteration, it directly overlays the undersampled true data $\mathbf{b}$ onto the $k$-space of the reconstructed image, while the un-sampled regions still adopt the values obtained through the projection step. Theoretically, only in the absence of noise, this substitution with the DC layer can guarantee the convergence of the iterative algorithm, which in practice is usually unattainable.
The ISTA algorithm for solving \eqref{eq:rec} can be formulated as,
\begin{equation}
	\footnotesize
	\label{eq:ista}
	\begin{cases}
		&\bar{\mathcal{X}} = \mathcal{X}^{n} - \mu \nabla_{\mathcal{X}^{n}} \left[ \frac{1}{2} \left\| \mathsf{A} (\mathcal{X}) - \mathbf{b} \right\|_2^2 \right] = \mathcal{X}^{n} - \mu \mathsf{A}^H\left[\mathsf{A}(\mathcal{X}^{n})-\mathbf{b}\right] \\
		&\mathcal{X}^{(k+1)} = \arg\min_{\mathcal{X}} \frac{1}{2\mu} \left\| \mathcal{X} - \bar{\mathcal{X}} \right\|_2^2 + \lambda \mathsf{\Phi}(\mathcal{X})=\operatorname{prox}_{\mu\lambda \mathsf{\Phi}}(\bar{\mathcal{X}}),
	\end{cases}
\end{equation}
where the operator $\operatorname{prox}_{\mu\lambda \mathsf{\Phi}}(\cdot)$ is the projection onto the set defined by the regularization $\Phi$ with the threshold $\mu\lambda$, with $\mu$ being the step size.

\subsection{Unrolling Networks using Composite Priors}
Composite priors-driven DUNs in dynamic MRI reconstruction mainly lie in the intersection of the low-rank and sparse priors. LplusS-Net \cite{huang2021deep} and SLR-Net \cite{ref_slrnet} represent two different types, where the former follows a low-rank plus sparse scheme, while the latter adopts a joint low-rank and sparse formulation which is the focus of this paper. Additionally, SOUL-Net \cite{chen2023soul} utilizes similar structures to SLR-Net but is applied in CT reconstruction. JSLR-Net \cite{zhong2024jslrnet} also follows the joint low-rank and sparse scheme, but it is unfolded by the HQS algorithm, which may need conjugate gradient (CG) iterations to solve the subproblems. However, CG iterations will increase the complexity of the network and make the training process more difficult \cite{aggarwal2018modl}.

In SLR-Net, the low-rank prior is leveraged through the Casorati matrix. The Casorati matrix is obtained by unfolding the dynamic MRI tensor $\mathcal{X}$ along the temporal dimension, resulting in an $HW \times T$ matrix. The nuclear norm of this matrix is then utilized to explore the correlation of temporal evolution curves. However, this approach neglects spatial correlations, and the process of unfolding a high-dimensional tensor into a matrix inevitably loses the high-dimensional structural features of the data. Therefore, we believe a superior approach would be to employ tensor low-rank priors to construct the network.

% Furthermore, both the sparse and low-rank constraints are imposed directly on the image, making them inseparable. Thus, solving the reconstruction model becomes more challenging, which consequently leads to a more complex network structure. Therefore, this motivates us to propose a simple yet efficient structure for the joint low-rank and sparse unrolling network.
% 然而，SLR-Net中的稀疏和低秩约束都是对图像本身施加的，因此不可分离，这也导致其求解重建模型较为复杂。不过，在过去基于优化的模型的研究中已经表明，联合低秩和稀疏模型与低秩plus稀疏的模型对图像的理解不同，而且基于优化的联合方法也取得了非常好的效果。

Furthermore, the joint low-rank and sparse model can be formulated as,
\begin{equation}
    \label{eq:opt}
    \min_{\mathcal{X}} \frac12 \| \mathsf{A}(\mathcal{X}) - \mathbf{b} \|_2^2 + \lambda_1 \mathsf{R}(\mathcal{X}) + \lambda_2 \mathsf{S}(\mathcal{X}),
\end{equation}
where $\mathsf{R}(\mathcal{X})$ and $\mathsf{S}(\mathcal{X})$ denote the LR and sparse priors, respectively. $\lambda_1$ and $\lambda_2$ are the balancing parameters.
SLR-Net has developed an algorithm that embeds ISTA within ADMM. Specifically, using ADMM and introducing auxiliary variables $\mathcal{T}$, the optimization problem \eqref{eq:opt} can be reformulated as,
\begin{equation}
    \footnotesize
    \label{eq:opt_admm}
    \begin{cases}
        &\min_{\mathcal{X}} \frac12 \| \mathsf{A}(\mathcal{X}) - \mathbf{b} \|_2^2  + \frac\rho2 \|\mathcal{X} - \mathcal{T}+\mathcal{L}\|_F^2  + \lambda_2 \mathsf{S}(\mathcal{X})\\
        &\min_{\mathcal{T}} \frac\rho2 \|\mathcal{X} - \mathcal{T}+\mathcal{L}\|_F^2 + \lambda_1 \mathsf{R}(\mathcal{T})\\
        &\mathcal{L} = \mathcal{L} + \eta(\mathcal{X} - \mathcal{T}),
    \end{cases}
\end{equation}
where $\mathcal{L}$ is the Lagrangian multiplier. ISTA is then embedded to solve the $\mathcal{X}$ subproblem, resulting in, 
\begin{equation}
    \footnotesize
    \label{eq:opt_pgdadmm}
    \begin{cases}
        & \begin{cases}
            & \mathcal{Z} = \mathcal{X} - \mu \nabla_\mathcal{X}\left[\frac12 \| \mathsf{A}(\mathcal{X}) - \mathbf{b} \|_2^2  + \frac\rho2 \|\mathcal{X} - \mathcal{T}+\mathcal{L}\|_F^2\right] \\
            & \min_{\mathcal{X}} \frac1{2\mu} \|  \mathcal{X} - \mathcal{Z} \|_2^2 + \lambda_2 \mathsf{S}(\mathcal{X})
            \end{cases} \\
        &\min_{\mathcal{T}} \frac\rho2 \|\mathcal{X} - \mathcal{T}+\mathcal{L}\|_F^2 + \lambda_1 \mathsf{R}(\mathcal{T})\\
        &\mathcal{L} = \mathcal{L} + \eta(\mathcal{X} - \mathcal{T}).
    \end{cases}
\end{equation}
Note that we have omitted the indexing notation for iterations for the sake of brevity. 
% Theoretically, this algorithm requires subsequent iterations to be computed after the convergence of PGD within the inner loop. However, in practice, deploying this algorithm as a deep unrolling network is impractical. SLR-Net adopts an approximate method by fixing the number of PGD iterations within the inner loop to 1. Actually, the model can be solved using ADMM alone, like what has been proposed in k-t SLR. The reason for embedding PGD lies in bypassing the conjugate gradient step in ADMM, thus reducing computational overhead.
% 从上式可知，三个临时变量$\mathcal{X}$, $\mathcal{T}$, $\mathcal{L}$的交互较为复杂，进而使得展开后的网络结构复杂，使得反向传播变得不高效。我们相信，更为直接且简单的迭代求解算法能够得到更加高效的展开网络结构，进而能够取得更好的重建效果。
From the above equation, it is evident that the interactions among the three temporary variables, $\mathcal{Z}$, $\mathcal{T}$, and $\mathcal{L}$, are relatively complex. This complexity renders the design of composite-prior DUNs difficult and cumbersome, while also hindering efficient backpropagation to some extent, consequently leading to suboptimal reconstruction results. We believe that a more simple iterative solving algorithm can also lead to a more efficient unrolling network structure and, consequently, achieving better reconstruction results and simplifying network design.

\section{Method}
\subsection{Composite Splitting Algorithm for Joint Low-Rank and Sparse Model}
The dynamic MRI image is a 3-way tensor and exhibits high-dimensional data structures and correlations. Therefore, we construct the low-rank and sparse regularizations from the perspective of tensor.

Specifically, the tensor low-rank regularization is constructed based on the transformed tensor nuclear norm (TTNN) \cite{zhang2024t2lr} under the framework of t-SVD \cite{ref_tsvd}, which is defined 
\begin{equation}
	\label{eq:lr}
	\mathsf{R}(\mathcal{X}) = \sum_{i=1}^{T}\|\mathsf{T}(\mathcal{X})^{(i)}\|_{*}.
\end{equation}
In the above equation, the operator $\|\cdot\|_*$ denotes the nuclear norm of a matrix, which is the sum of its singular values, and $\mathsf{T}$ is a CNN-learned transformation that maps the input tensor to a low-rank domain. The subscript $(i)$ denotes the $i$-th frontal slice of the tensor, i.e., for a tensor $\mathcal{Y}=\mathsf{T}(\mathcal{X})$, $\mathcal{Y}^{(i)}=\mathcal{Y}(:,:,i), i=1,2,...,T$ holds. 

The TTNN can be interpreted as the sum of the nuclear norms of the frontal slices in the transformed domain. The transformation $\mathsf{T}$ applied to the entire tensor allows for the extraction of high-dimensional structural information. Consequently, within the transformed domain, any frontal slice may encompass all the information from the original image domain. While we aim to provide a comprehensive interpretation of TTNN, it is essential to note that TTNN possesses a complete and rigorous mathematical definition and derivation (see \cite{zhang2024t2lr}).

The tensor sparse regularization is constructed based on the tensor $l_1$ norm in the CNN-learned domain, which is defined as,
\begin{equation}
	\label{eq:sp}
	\mathsf{S}(\mathcal{X}) = \|\mathsf{D}(\mathcal{X})\|_1,
\end{equation}
where $\mathsf{D}$ is a CNN-learned transformation that maps the input tensor to a sparse domain. The tensor $l_1$ norm is the sum of the absolute values of all elements in the tensor. 

To efficiently solve the joint low-rank and sparse model \eqref{eq:opt}, the optimization problem \eqref{eq:opt} can be decomposed into the following subproblems using the composite splitting algorithm \cite{huang2011composite, combettes2008proximal},
\begin{align}
		&\bar{\mathcal{X}} = \mathcal{X}^{n}-\mu \mathsf{A}^H\left[\mathsf{A}(\mathcal{X}^{n})-\mathbf{b}\right], \\
		&\mathcal{Y}_1 = \arg\min_{\mathcal{X}} \frac{\omega_1}{2\mu} \|\mathcal{X} - \bar{\mathcal{X}}\|_F^2 + \lambda_1 \mathsf{R}(\mathcal{X}),\\
		&\mathcal{Y}_2 = \arg\min_{\mathcal{X}} \frac{\omega_2}{2\mu} \|\mathcal{X} - \bar{\mathcal{X}}\|_F^2 + \lambda_2 \mathsf{S}(\mathcal{X}),\\
		&\mathcal{X}^{n+1} = \omega_1\mathcal{Y}_1 + \omega_2\mathcal{Y}_2, \label{eq:combine}
\end{align}
where the first step is the gradient descent, the second and the third steps are the projection steps, and $\omega_1$ and $\omega_2$ with $\omega_1+\omega_2$=1 are the balancing parameters. 

From \eqref{eq:lr} and the corresponding transformed tensor singular value thresholding algorithm \cite{zhang2024t2lr}, the $\mathcal{Y}_1$ subproblem can be solved as,
\begin{equation}
	\label{eq:lr_pgd}
	\mathcal{Y}_1 = \mathsf{T}^H\circ\operatorname{SVT}_{\frac{\mu\lambda_1}{\omega_1}}\circ\mathsf{T}(\bar{\mathcal{X}}),
\end{equation}
where $\operatorname{SVT}_{\tau}(\cdot)$ is the singular value thresholding operator with threshold $\tau$ for each frontal slice of the tensor and $\mathsf{T}^H$ is the adjoint transformation of $\mathsf{T}$. We further employ a widely used trick \cite{ref_slrnet} to allocate a distinct threshold to each frontal slice individually (more details in Section \ref{sec:lr}).

The $\mathcal{Y}_2$ subproblem can be solved by the soft thresholding operator \cite{donoho1995noising} in the transformed domain, i.e.,
\begin{equation}
	\label{eq:sp_pgd}
	\mathcal{Y}_2 = \mathsf{D}^H\circ\operatorname{ST}_{\frac{\mu\lambda_2}{\omega_2}}\circ\mathsf{D}(\bar{\mathcal{X}}),
\end{equation}
where $\operatorname{ST}_{\tau}(\cdot)$ is the soft thresholding operator with threshold $\tau$ for each element of the tensor. It is worth noting that the ST operator only uses a single threshold for all elements in the tensor, which may limit the flexibility of the sparse constraint. We propose a novel attention-based soft thresholding operator in Section \ref{sec:s} to address this issue.

% 从上述算法可以看出，我们所设计的算法具有高效但简单的结构，中间变量X和Y仅在当前迭代中相互交互，而不会在迭代之间构建错综复杂的关系，不像SLR-Net中的算法。同时，我们的算法可以视作ISTA的合理延伸，其类似两个基于R或S的但约束ISTA的交替，只是在最后一个迭代步增加了两个约束映射的线性组合。因此，在ISTA经过深度展开网络社区的广泛认可和应用的基础上，我们的算法很好的继承了ISTA对于展开网路的诸多优点。
From the above algorithm, it can be observed that our designed algorithm possesses an efficient yet simple structure. Intermediate variables $\bar{\mathcal{X}}$ and $\mathcal{Y}$ interact only within the current iteration, avoiding intricate relationships between iterations, unlike the algorithm \eqref{eq:opt_pgdadmm} in SLR-Net \cite{ref_slrnet}. Moreover, our algorithm can be considered a reasonable extension of ISTA, exhibiting a similar alternation of two ISTA ($\mathsf{R}(\mathcal{X})$ and $\mathsf{S}(\mathcal{X})$-constrained ISTAs) with the addition of a linear combination of two proximal mappings in the last iteration step \eqref{eq:combine}. Therefore, building upon the widespread recognition and application of ISTA in the DUN community, our algorithm also inherits the advantages of ISTA for DUNs.

% 另外，类似与ISTA可以结合Nesterov加速步来提高收敛速度，并且FISTA-Net也已经证明增加加速步可以提高重建效果。因此，受此启发，我们也可以在我们的算法中增加Nesterov加速步，进而提高重建效果。
Additionally, similar to ISTA's ability to incorporate the Nesterov acceleration step for improved convergence speed, and as demonstrated by FISTA-Net \cite{xiang2021fista}, adding the acceleration step can enhance reconstruction effectiveness. Thus, inspired by these findings, we also integrate Nesterov acceleration steps into our algorithm. In summary, our algorithm is depicted in Alg.\ref{algo_cps}.

\begin{algorithm}[htbp]
	\SetKwData{Left}{left}
	\SetKwData{This}{this}
	\SetKwData{Up}{up} 
	\SetKwFunction{Union}{Union}
	\SetKwFunction{FindCompress}{FindCompress} 
	\SetKwInOut{Require}{Require}
	\SetKwInOut{Initialize}{Initialize}
	\SetKwInOut{Return}{Return}
	
	\Require{$\{\mathsf{T},\mathsf{T}^H,\mathsf{D},\mathsf{D}^H,\mathbf{\mu},\lambda_1,\lambda_2,\omega_1,\omega_2\}$} 
	\Initialize{$\mathcal{X}^0 = \mathsf{A}^H\mathbf{b}$, $t^0=1$}
	 \BlankLine 
	 
	%  \emph{special treatment of the first line}\; 
	 \For{$n=1,\dots$, until convergence}{ 
		$\bar{\mathcal{X}} = \mathcal{X}^{n}-\mu \mathsf{A}^H\left[\mathsf{A}(\mathcal{X}^{n})-\mathbf{b}\right]$ \;
		$\mathcal{Y}_1 = \mathsf{T}^H\circ\operatorname{SVT}_{\frac{\mu\lambda_1}{\omega_1}}\circ\mathsf{T}(\bar{\mathcal{X}})$\;
		$\mathcal{Y}_2 = \mathsf{D}^H\circ\operatorname{ST}_{\frac{\mu\lambda_2}{\omega_2}}\circ\mathsf{D}(\bar{\mathcal{X}})$\;
		$\mathcal{Z}^{n} = \omega_1\mathcal{Y}_1 + \omega_2\mathcal{Y}_2$\;
    	\quad \\
		\emph{Acceleration Step}:\\
		$t^{n+1}=\frac{1+\sqrt{1+4(t^n)^2}}{2}$\;
		$\mathcal{X}^{n+1} = \mathcal{Z}^n+\frac{t^n-1}{t^{n+1}}(\mathcal{Z}^n-\mathcal{Z}^{n-1})$
 	 } 
	  \BlankLine 
      \caption{Composite Splitting Algorithm for Joint Low-Rank and Sparse Model}
 	 	  \label{algo_cps} 
	\end{algorithm}

\subsection{JotlasNet}
\label{sec:net}
% 从迭代优化算法的角度出发，算法1中的“require”项含有大量的需要预定义的变量，包括低秩和稀疏变换域以及各种平衡参数。然而，一方面，变换域的选择与重建精度之间没有直接的关系，而且，平衡参数的选择往往也是通过经验进行确定，顶多是通过小数量的全采样MRI图像进行确定，缺乏鲁棒性，而且人工选择这些参数尝尝是极为耗时的。展开为深度展开网络能够将算法1中的“require”项全部变成可学习参数，进而通过监督机制构建起参数选择和重建精度之间的联系，从而能够取得更好的重建效果。
% From the perspective of iterative optimization algorithms, the "Require" section in Algorithm \ref{algo_cps} contains numerous variables that need to be pre-defined, including the low-rank and sparse transformed domains, as well as various balancing parameters. However, on the one hand, the choice of the transformed domain does not have a direct relationship with the reconstruction accuracy. Moreover, the selection of balancing parameters is often determined empirically, typically through a small number of fully sampled MRI images, lacking robustness. Additionally, manually selecting these parameters can be highly time-consuming. Transforming this into a deep unrolling network allows us to convert all the variables in the "Require" section of Algorithm \ref{algo_cps} into learnable parameters. This, in turn, establishes a connection between parameter selection and reconstruction accuracy through a supervised mechanism, leading to improved reconstruction results. 

\begin{figure*}
	\centering
	\includegraphics[width=1\textwidth]{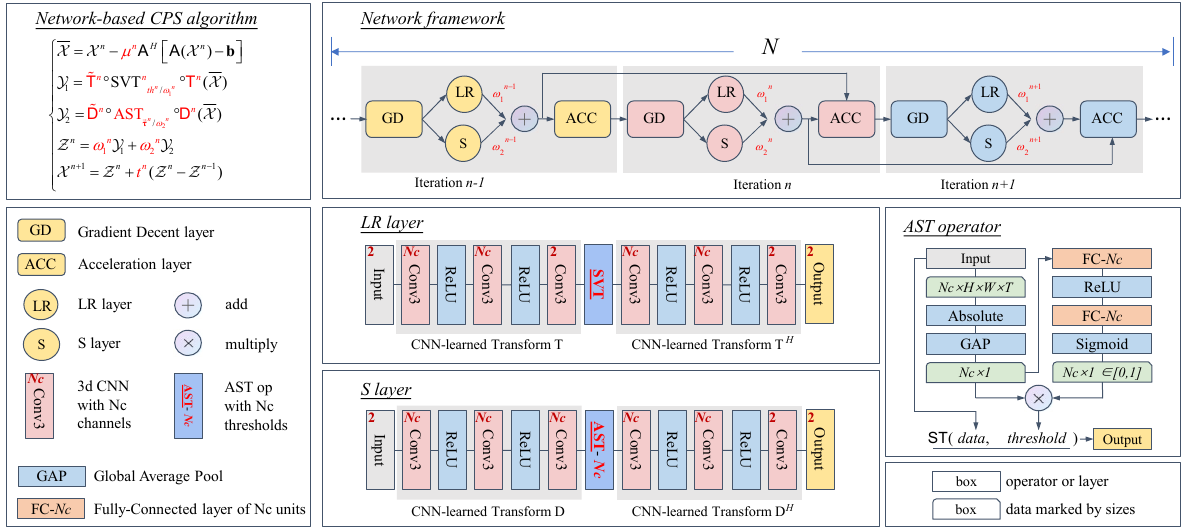}
	\caption{The proposed JotlasNet. The network is unfolded from \eqref{eq:net}. The red-marked symbols are learnable parameters. Five layers are included: Gradient Descent (GD), Low-Rank (LR), Sparse (S), combination, and Acceleration (ACC) layers. The LR and S layers are detailed in the text.}
	\label{fig:network}
\end{figure*}

Unfolding Alg.\ref{algo_cps} into a DUN allows for transformations and hyperparameters therein becoming neural networks and learnable parameters. Therefore, we can also further enhance the flexibility by assigning a unique set of parameters to each iteration. Finally, we propose the JotlasNet as shown in Fig.\ref{fig:network}, which is unfolded from the following algorithm.
\begin{equation}
	\label{eq:net}
	\begin{cases}
		\bar{\mathcal{X}} &= \mathcal{X}^{n}-{\mu^n} \mathsf{A}^H\left[\mathsf{A}(\mathcal{X}^{n})-\mathbf{b}\right] \\
		\mathcal{Y}_1 &= {\tilde{\mathsf{T}}^n}\circ\operatorname{SVT}_{{th^n}/{\omega_1^n}}\circ{\mathsf{T}^n}(\bar{\mathcal{X}})\\
		\mathcal{Y}_2 &= {\tilde{\mathsf{D}}^n}\circ{\operatorname{AST}}_{{\vec{\mathbf{\tau}}^n}/{\omega_2^n}}\circ{\mathsf{D}^n}(\bar{\mathcal{X}})\\
		\mathcal{Z}^{n} &= {\omega_1^n}\mathcal{Y}_1 + {\omega_2^n}\mathcal{Y}_2\\
		\mathcal{X}^{n+1} &= \mathcal{Z}^n+{t^n}(\mathcal{Z}^n-\mathcal{Z}^{n-1})
		\end{cases},
\end{equation}
where we replace $\mu\lambda_1$ and $\frac{t^n-1}{t^{n+1}}$ with a single learnable parameter $th$ and $t$, respectively. The ST operator with a single threshold $\mu\lambda_2$ is replaced by the proposed AST operator with a vector of thresholds $\vec{\mathbf{\tau}}$ for each channel of the input data. The red-marked symbols in Fig.\ref{fig:network} are learnable and the superscript $n$ denotes they are unique for each iteration.

% 除了LR和S layers，其他的层都未嵌入例如CNN等神经网络，完全按照迭代算法的公式在深度学习框架中计算，只是标红的参数设置为可学习参数。值得注意的是，\mu>0, w1+w2=1, 0<+t<=1。我们可以通过relu，softmax，和sigmoid来相对应的对参数进行合理约束。
% 我们将在下面对LR和S layers进行详细的介绍。
The JotlasNet consists of five layers, the Gradient Descent (GD), Low-Rank (LR), Sparse (S), combination and Acceleration (ACC) layers, corresponding to the five steps of \eqref{eq:net}, respectively. 
Except for the LR and S layers, the remaining layers are not incorporated with neural networks, such as CNN. Instead, they are computed in the deep learning framework according to the iterative algorithm's formulas, with only the parameters becoming trainable. It is worth noting that $\mu>0$, $\omega_1+\omega_2=1$, and $t\in[0,1]$. We can apply ReLU, softmax, and sigmoid functions to reasonably constrain the parameters accordingly. The LR and S layers are described in detail as follows.

\subsubsection{The LR Layer}
\label{sec:lr}
% 该layer对应于公式(16)的Y1 step，起源于Algorithm 1的Y1 step。得益于展开网络的架构，我们用两个独立的CNN来学习Algorithm 1中的TH和T两个变换，并将其标记为Tilde T和T。这两个CNN具有相同的结构，即具有三个3d卷积层，卷积核的大小均为3x3x3，步长为1，padding为1，输出通道数分别为Nc，Nc，和2。最后一个卷积层的通道数和输入输出通道数为2的原因是我们将MRI复数数据分为实部和虚部两通道。两个ReLU激活函数穿插在三个卷积层之间，最后一个卷积层不激活从而保证输出值的负数部分不会被截断。在CNN变换域内，我们将双通道实数数据合成为复数数据。
This layer corresponds to the $\mathcal{Y}_1$ step in Eq.\eqref{eq:net}. We utilize two independent CNNs to learn the transformations $\mathsf{T}^H$ and $\mathsf{T}$ in Alg.\ref{algo_cps}, denoted as $\tilde{\mathsf{T}}$ and $\mathsf{T}$, respectively. These two CNNs share the same structure, consisting of three 3D convolutional layers with kernel sizes of $3\times 3 \times 3$ and the stride of 1. The output channels for the three convolutional layers are $Nc$, $Nc$, and 2, respectively. The reason for having 2 channels in the last convolutional layer is that we split the MRI complex data into real and imaginary parts. Two ReLU activation functions are interspersed between the three convolutional layers, with the last convolutional layer not being activated to ensure that negative values are not truncated in the output. In the CNN transformation domain, we synthesize the dual-channel real data into complex data, which we denote as $\mathcal{F}_1$. The learned SVT operator in the middle is applied to each frontal slice of $\mathcal{F}_1$. For the $i$-th frontal slice with its singular value decomposition (SVD) as $\mathcal{F}_1^{(i)}=\mathbf{U}\mathbf{\Sigma}\mathbf{V}^H$, the learned SVT operates as follows.
\begin{equation}
	\footnotesize
		\operatorname{SVT}_{th/\omega_1}(\mathcal{F}_1^{(i)})= \mathbf{U} \cdot \operatorname{ReLU}\left(\mathbf{\Sigma} - \frac{\operatorname{sigmoid}(th)\cdot\sigma_{max}}{\omega_1}\right) \cdot \mathbf{V}^H 
  \end{equation}
where we omit the superscript $n$ for brevity, $\sigma_{max}$ is the maximum singular value of $\mathcal{F}_1^{(i)}$, and $\operatorname{sigmoid}(\cdot)$ is the sigmoid function. 

% 值得注意的是，在展开的过程中，我们放松了Algorithm 1中对于TH和T为共轭的约束，让两个CNN自由地根据数据学习。这样可以同时利用显式的低秩先验和CNN提取的隐式图像先验信息，从而能够增强重建效果。具体来说，如果th在训练的过程中被学习成0，或者一个很小的值，那么该LR layer将变成一个纯CNN层，从而能够学习到超出低秩的其他重建所需要的先验信息。
It is worth noting that during the unrolling process, we relax the constraint in Alg.\ref{algo_cps} that $\mathsf{T}^H$ and $\mathsf{T}$ must be adjoint, allowing the two CNNs to learn freely from the data. This approach enables us to simultaneously utilize explicit low-rank priors and implicit image prior information extracted by CNNs, thereby enhancing the reconstruction performance \cite{zhang2024t2lr}. Specifically, if the threshold $th$ is learned to be close to zero or a very small value during training, the LR layer will effectively become a pure CNN layer, capable of learning additional prior information beyond low-rank constraints that may be required for reconstruction.

\subsubsection{The S Layer}
\label{sec:s}
% 类似于LR layer，该layer对应于公式(17)的Y2 step，起源于Algorithm 1的Y2 step。我们同样用两个独立的CNN来学习Algorithm 1中的$\vec{\mathbf{\tau}}$和$\mathsf{D}$两个变换，并将其标记为$\tilde{\mathsf{D}}$和$\mathsf{D}$。与LR layer不同的是，用于学习$\tilde{\mathsf{D}}$的第三个卷积层的输出通道数被设置成$Nc$增大数据维度以增加冗余性，从而能够提取到更为丰富的稀疏先验信息。这主要是得益于稀疏约束无需进行奇异值分解从而具有较低的计算复杂度。同时，在CNN变换域内，我们也不在将双通道实数数据合成为复数数据，而是直接对每个通道进行稀疏约束。
Similar to the LR layer, this layer corresponds to the $\mathcal{Y}_2$ step in Eq.\eqref{eq:net}. Similarly, we use two independent CNNs to learn the transformations $\mathsf{D}^H$ and $\mathsf{D}$ in Alg.\ref{algo_cps}, denoted as $\tilde{\mathsf{D}}$ and $\mathsf{D}$, respectively. Unlike the LR layer, the output channels of the third convolutional layer used to learn $\tilde{\mathsf{D}}$ are set to $Nc$ to increase data dimensionality and redundancy, thereby extracting richer sparse prior information. This is mainly due to the sparse constraint not requiring SVD, resulting in lower computational complexity. Additionally, in the CNN transform domain, we do not synthesize dual-channel real data into complex data; instead, we directly impose sparse constraints on each channel.

% 目前的展开网络对于稀疏约束是通过软阈值算子来实现的。其中，仅利用一个阈值对所有通道的数据进行约束。我们相信，这种方式会限制稀疏约束的表达能力，因为不同通道的数据可能具有不同的稀疏性。因此，启发于[1]，我们提出了一种新的稀疏约束算子，即自适应稀疏阈值算子（AST）。具体来说，对于一个$Nc\times H\times W \times T$的输入数据 $\mathcal{F}_2$，AST首先对其取绝对值并进行全局平均池化，得到一个$Nc \times 1$的向量$\vec{\mathbf{f}}$，然后利用被ReLU分离的两层全连接层进行学习，并通过sigmoid函数得到一个自注意力权重向量$\vec{\mathbf{w}}$。最后，将$\vec{\mathbf{w}}$与$\vec{\mathbf{f}}$逐元素相乘，得到一个$Nc \times 1$的自适应稀疏阈值向量$\vec{\mathbf{\tau}}$。具体来说，对于第i个通道的数据$\mathcal{F}_2^{[i]}$，AST的操作如下。
Current DUNs implement sparse constraints using an ST operator, where only one threshold is used to constrain data across all channels. We believe that this approach may limit the expressive power of sparse constraints, as data in different channels may exhibit different sparsity patterns. Therefore, inspired by \cite{zhao2019deep}, we propose a novel Attention-based Soft Thresholding operator (AST), assigning a learnable threshold to each channel of the data in the CNN-learned sparse domain. The diagram of AST is shown in Fig. \ref{fig:network}. Specifically, for an input data $\mathcal{F}_2$ of size $Nc \times H \times W \times T$, AST first takes the absolute value and performs global average pooling (GAP) to obtain a vector $\vec{\mathbf{f}}$ of size $Nc \times 1$. Then, two fully connected layers separated by ReLU activation and later the sigmoid function are utilized to obtain a self-attention weight vector $\vec{\mathbf{w}}$. Element-wise multiplication of $\vec{\mathbf{w}}$ and $\vec{\mathbf{f}}$ yields an adaptive sparse threshold vector $\vec{\mathbf{\tau}}=[\tau_1, \tau_2, \cdots, \tau_{Nc}]^T$. Finally, the ST operator with the threshold $\tau_i$ is applied to the $i$-th channel of the data $\mathcal{F}_2^{[i]}$ as follows.
\begin{equation}
	\operatorname{ST}_{\tau_i/\omega_2}(\mathcal{F}_2^{[i]})=\frac{\mathcal{F}_2^{[i]}}{|\mathcal{F}_2^{[i]}|}\cdot\operatorname{ReLU}\left(|\mathcal{F}_2^{[i]}|-\frac{\tau_i}{\omega_2}\right)
\end{equation}
where $|\mathcal{F}_2^{[i]}|$ denotes the absolute value of $\mathcal{F}_2^{[i]}$ and $\frac{\mathcal{F}_2^{[i]}}{|\mathcal{F}_2^{[i]}|}$ is the element-wise divide operation, i.e., the sign function of $\mathcal{F}_2^{[i]}$.

\subsubsection{Loss Function}
% 我们的目标是最小化重建误差，因此我们的损失函数设置为均方误差。具体来说，我们的损失函数为
We utilize the mean squared error (MSE) as the loss function, i.e.,
\begin{equation}
	\label{loss_func}
		Loss = \sum_{(\mathcal{X}_{GT}, \mathbf{b}) \in \Omega}\Vert \mathcal{X}_{GT} - f_{net}(\mathbf{b}|\theta) \Vert_F^2.
  \end{equation}
where $\Omega$ is the training set, $\mathcal{X}_{GT}$ is the ground truth, $f_{net}(\mathbf{b}|\theta)$ is the output of the network with parameters $\theta=\{\mathsf{T}^n,\tilde{\mathsf{T}}^n,\mathsf{D}^n,\mathsf{D}^n,\mathbf{\mu}^n,th^n,\vec{\mathbf{\tau}}^n,\omega_1^n,\omega_2^n, FC^n | n=1\cdots N\}$, and $FC$ denotes the two fully connected layers in the AST operator.

\subsubsection{Implementation Details}
% 出于对计算负载和重建精度的折中考虑，我们设置JotlasNet的迭代数N为15。卷积层和全连接层的通道数和隐藏单元数Nc设置为16。我们使用Adam优化器进行训练，其参数beta1=0.9，beta2=0.999，epsilo=10-8，学习率设置为0.001并且以0.95的衰减率进行衰减。我们使用TensorFlow框架进行实验，batch size设置为1。
Considering the trade-off between computational burden and reconstruction accuracy, we set the number of iterations \( (N) \) for JotlasNet to 15. The number of channels for convolutional layers or hidden units for fully connected layers \( (N_c) \) is set to 16, resulting in a total of about 708k parameters. We train the network using the Adam optimizer \cite{kingma2014adam} with parameters \( \beta_1 = 0.9 \), \( \beta_2 = 0.999 \), \( \epsilon = 10^{-8} \), and an initial learning rate of 0.001, decayed by a factor of 0.95 \cite{zeiler2012adadelta}. The model is implemented using the TensorFlow framework \cite{abadi2016tensorflow} with a batch size of 1.

\subsubsection{Computational Complexity}
Suppose that the input of JotlasNet $\mathcal{X}$ has the size of $H\times W\times T$ with $H > W$, and the number of coils is denoted as $C$. The GD layer only consists of linear operations and repeated fast Fourier transforms on $H$ and $W$ spatial dimensions across temporal dimension $T$ and coil channel $C$, resulting in a complexity of $\mathcal{O}(HW\log(HW)TC)$. For the S layer, the total complexity of the two CNNs (the transform $\mathsf{D}$ and its transpose) is $HWT\times 3^3 \times 16 \times (16+16+2) \times 2$, i.e., $\mathcal{O}(HWT)$. Here, we directly use the kernel size and the number of channels from the implementation details to simplify the calculation. The cost of the AST operator is dominated by linear operations, leading to a complexity of $\mathcal{O}(HWT)$. Note that the calculation of the attention maps depends on the number of channels, which is negligible compared to the linear operations. For the LR layer, the two CNNs have the same complexity as the S layer, resulting in $\mathcal{O}(HWT)$. The SVT operator is dominated by the tensor SVD, whose complexity is $\mathcal{O}(HW^2T)$. The ACC layer only involves linear operations, resulting in a complexity of $\mathcal{O}(HWT)$. Therefore, for $N$ iterations, the total complexity of JotlasNet is $\mathcal{O}(NHWT(\log(HW)C+W))$.

\section{Experiments and Results}
\subsection{Datasets and Experimental Setup}
% 我们在两个公开的心脏电影数据集OCMR、CMRxRecon上进行了充足的实验，证明了我们所提出JotlasNet的有效性。
We conducted experiments on two publicly available cardiac cine MRI datasets, OCMR and CMRxRecon. The specific datasets and experimental configurations are listed in Tab.\ref{tab:setup}.

% \begin{table}[htbp]
% 	\centering
% 	\caption{Datasets and Experimental Setup}
% 	  \begin{tabular}{lll}
% 	  \toprule
% 	  dataset & coil  & sampling \\
% 	  \midrule
% 	  \multicolumn{1}{c}{\multirow{2}[1]{*}{{OCMR}}} & single-coil & radial-16,30 ; VDS-8,10 \\
% 			& multi-coil & VSITA-8,12,24 \\
% 	  {CMRxRecon} & multi-coil & equispaced-4,8,10 \\
% 	  \bottomrule
% 	  \end{tabular}%
% 	\label{tab:setup}%
%   \end{table}%
  
\begin{table}[htbp]
	\centering
	\caption{Datasets and Experimental Setup}
	\resizebox{\linewidth}{!}{\begin{tabular}{cccc}
	  \toprule
	  dataset & coil  & sampling & section \\
	  \midrule
	  \multirow{2}[1]{*}{\textbf{OCMR}} & single-coil & radial-16,30 ; VDS-8,10 & \ref{sec:sigcoil}, \ref{sec:ablation} \\
			& multi-coil & VSITA-8,12,24 & \ref{sec:multicoil} \\
	  \textbf{CMRxRecon} & multi-coil & equispaced-4,,8,10 & \ref{sec:multicoil} \\
	  \bottomrule
	  \end{tabular}%
	  \label{tab:setup}}%   
  \end{table}%

\subsubsection{OCMR Dataset}
This dataset \cite{ref_ocmr} consists of 204 fully sampled cardiac cine MRI raw data, which were acquired from 74 subjects collected on three Siemens MAGNETOM scanners: Prisma (3T), Avanto (1.5T), and Sola (1.5T). Both the short-axis and long-axis views are included. For our experiments, we allocated 124 data for training, 40 for validation, and 40 for testing. For the training set, considering the limited data quantity, we applied appropriate data augmentation techniques. We cropped the data to a size of $128\times128\times16$ based on the dimensions of all data in the training set, with strides of 32, 32, and 8 in the spatial (height, width) and temporal directions, respectively. Finally, a total of 1848 fully sampled training samples were obtained.

% 我们在该数据集上进行了单线圈和多线圈的实验。其中，单线圈实验中，我们利用ESPIRiT估计的线圈敏感度将所有多线圈的raw data进行了单线圈合成。我们探索了两种不同采样模板情况下的重建表现，并且没种采样模板采用了两种不同的加速倍数。具体来说，16和30条采样线的pusedo-radial采样模板和8倍和10倍加速的变密度随机采样（vds）被采用。在多线圈实验中，我们采用了VISTA采样模板，并且在8,12，和超高倍数24加速下对所提出JotlasNet进行了评估。
In our experiments on this dataset, both single-coil and multi-coil scenarios were considered. For the single-coil experiments, we synthesized single-coil data from the multi-coil raw data using the coil sensitivity maps (CSM) estimated by ESPIRiT \cite{ref_multicoil}. We explored the reconstruction performance under two different sampling masks, each with two different acceleration factors. Specifically, pseudo-radial sampling \cite{lingala2011accelerated} with 16 and 30 sampling lines and variable density random sampling (vds) with acceleration factors of 8 and 10, were adopted. 
The sampling lines in radial patterns are generated uniformly in the polar coordinate system at each time frame, while rotated by a small random angle across frames. The vds patterns undersample the grid by constructing a probability density function and randomly draw indices from that density to ensure undersampling less near the k-space origin and more in the periphery of k-space.
In these single-coil cases, we trained our JotlasNet and the comparative networks from scratch for 50 epochs. In the multi-coil experiments, we also used the ESPIRiT \cite{ref_multicoil} to estimate the CSMs for the network input, but we directly used the root sum of squares (RSS) images \cite{roemer1990nmr} as the ground truth for evaluation. The reason for this is that the estimated CSMs may not be accurate enough for the multi-coil data \cite{muckley2021results}. We employed the VISTA sampling mask \cite{ahmad2015variable} and evaluated the proposed JotlasNet under accelerations of 8, 12, and an ultra-high acceleration of 24. In VISTA-8, we trained the networks from scratch for 50 epochs, while finetuned them in other cases for 20 epochs using the trained weights in VISTA-8. The experiments in this dataset were conducted using one NVIDIA Quadro GV100 GPU with 32GB memory.

\subsubsection{CMRxRecon Dataset}
% 我们用了CMRxRecon中的120个cases的全采样raw data实验。这些数据是由一个3T Siemens MAGNETOM Vida scanner with a dedicated cardiac coil made up of 32 channels采集。每个case都包含6~12个slcies的数据，并且同时包含short-axis， two-chamber, three-chamber and four-chamber and long-axis views。我们按case分割为100,10,10作为训练，验证，和测试集。最终，我们得到1247,124,121个全采样的训练，验证，和测试raw data。由于该数据集具备较大体量的原始数据，同时，为了保证实验尽可能的贴近临床实际情况，我们不对数据集进行任何的诸如裁剪等数据增强。
We used fully sampled raw data from 120 cases in CMRxRecon \cite{wang2023cmrxrecon} for experiments. These data were acquired using a 3T Siemens MAGNETOM Vida scanner with a dedicated cardiac coil comprising 32 channels. Each case contains data from 6 to 12 slices, including short-axis, two-chamber, three-chamber, four-chamber, and long-axis views. We divided all the data cases into training, validation, and test sets at a ratio of 100:10:10. Consequently, we obtained 1247, 124, and 121 fully sampled training, validation, and test raw data samples/slices, respectively. % Given the substantial volume of raw data in this dataset and to ensure experiments closely resemble clinical scenarios, we refrained from any data augmentation techniques such as cropping. 

% 我们在CMRxRecon数据集上对JotlasNet的多线圈重建表现进行评估。我们采用仅对相位编码方向欠采样的pseudo-equispaced sampling masks with中心24个编码全采样。在不计算自校准信号的情况下，我们采用8,10，和12进行实验。
Only the multi-coil scenario was considered on the CMRxRecon dataset. We employed 1D pseudo-equispaced sampling masks \cite{knoll2020advancing,zbontar2018fastmri} (undersampling only in the phase-encoding direction). The central 24 columns were fully sampled as the auto-calibration signal region \cite{ref_multicoil}. The accelerations of 8, 10, and 12 (without considering auto-calibration signals) were used. We also used the ESPIRiT to estimate the CSMs for the network input and directly utilized the RSS images as the ground truth for evaluation. In the 8X case, we trained the networks from scratch for 50 epochs. In other scenarios, we finetuned them for 20 epochs using the weights trained in the 8X case. The experiments in this dataset were conducted using one NVIDIA A100 GPU with 80GB memory.

\subsubsection{Comparative Methods and Metrics}
We compared our proposed JotlasNet with four state-of-the-art (SOTA) DUNs for dynamic MRI reconstruction, including LplusS-Net \cite{huang2021deep}, ISTA-Net \cite{ref_ISTANET}, DCCNN \cite{ref_DCCNN} and SLR-Net \cite{ref_slrnet}. All comparative networks adopted the network architecture and parameter settings provided in their respective papers, and were implemented using their open-source code. The only difference lies in retraining them on the datasets we used. We also provided the zero-filled reconstruction results and the fully sampled ground truth as the baseline and label, respectively. We evaluated the reconstruction performance using the peak signal-to-noise ratio (PSNR) and structural similarity index measure (SSIM) as the metrics.

\subsection{Single-coil Results} 
\label{sec:sigcoil}
\begin{figure*}
	\centering
	\includegraphics[width=1\textwidth]{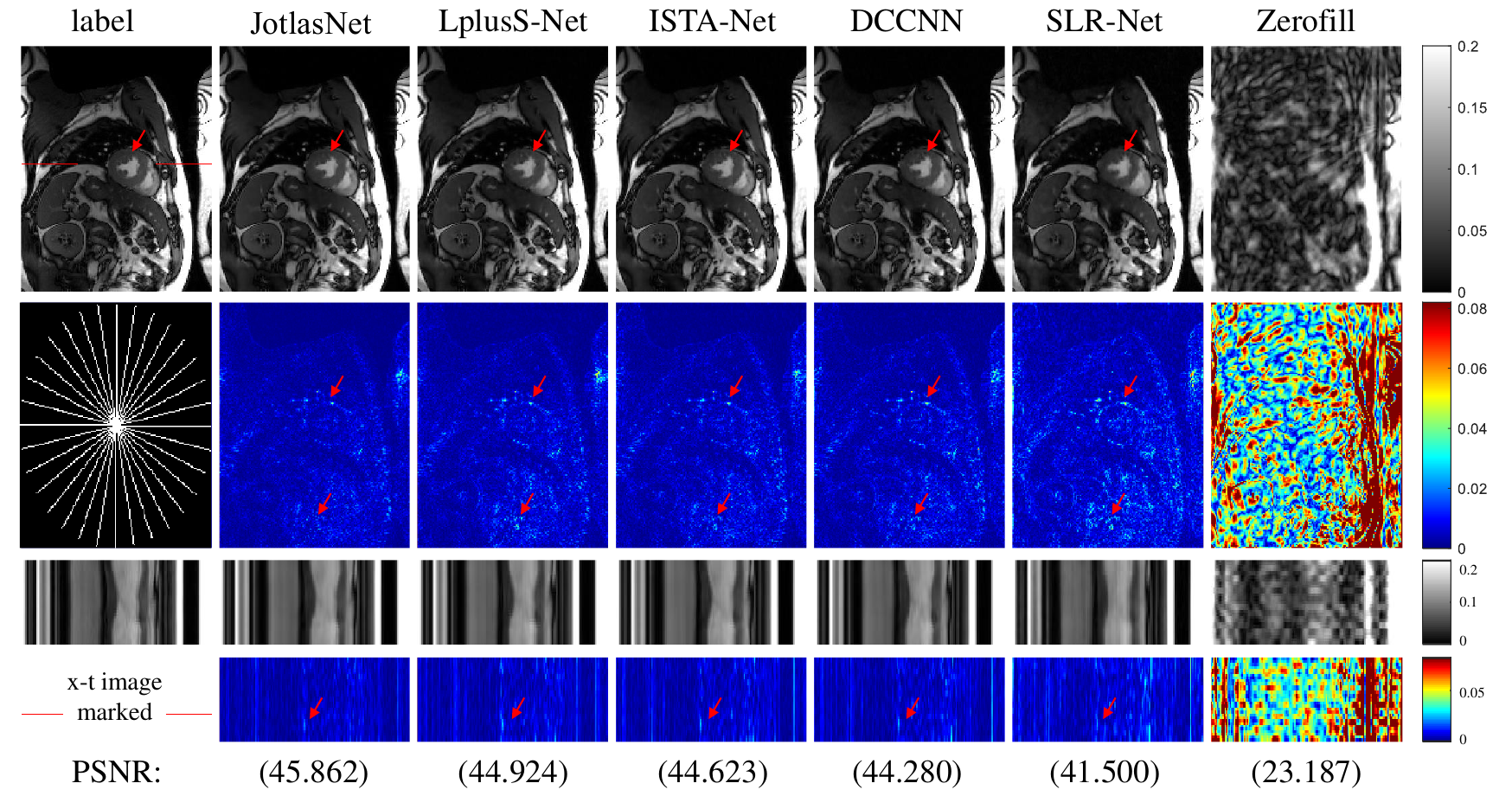}
	\caption{Reconstruction results on the OCMR dataset (single-coil) using radial sampling with 16 lines. The first row shows the reconstructed images at a specific time frame, and the second row shows the corresponding error maps, except for the first image depicted the sampling mask. The third and fourth rows show the x-t images and the corresponding error maps, respectively. The position of x-t images is marked with red lines on the label image. The PSNR values respected to this test image are also listed.}
	\label{fig:radial}
\end{figure*}

% 我们在表3中列出了各种重建方法在OCMR数据集(单线圈)上的重建结果。最好的结果用红色标出，第二好的结果用蓝色标出。从定量指标中可以发现，在这种情况下，我们的JotlasNet网络在所有实验的采样模板下都能够取得最好的PSNR和SSIM。另外，我们也给出了相应的测试时间，我们的网络能够取得可以比较的重建速度。稍长的时间主要是消耗在低秩layer中的奇异值分解的计算，因为我们知道奇异值分解的计算复杂度在n\times n矩阵上是O(n^3)，而卷积层的计算复杂度仅为O(n^2)。但是，从物理意义上来说，低秩先验的引入使得网络能够更好地利用数据的冗余性，从而能够更好地重建图像。在16条线的radial采样情况下的可视化结果如图2所示。我们在图中展示了测试集中的某个数据的重建图像，并且也同时展示了特定帧和x-t图像的重建结果。从图中可以看出，我们的JotlasNet网络能够重建出更清晰的纹理，同时也能够更好地保留细节。从误差图上可以看到，在组织较为复杂的区域且在运动较为剧烈的区域，我们的网络能够取得更好的重建效果。
In Tab.\ref{tab:sigcoil}, we present the quantitative results of various networks on the OCMR dataset (single-coil). The symbol '-' indicates that LplusS-Net cannot be trained on the corresponding sampling cases, which is due to the instability of SVD gradients \cite{wang2021robust} and the occurrence of NaN (not a number) values when computing specific matrix SVD in TensorFlow. From the quantitative metrics, it can be observed that JotlasNet achieves the highest PSNR and SSIM across all experiments under different sampling patterns. Additionally, we provide the corresponding inference times, demonstrating comparable reconstruction speeds achieved by our network. The slightly longer time is mainly attributed to the computation of SVD in the low-rank layers, since the computational complexity of SVD on an $n\times n$ matrix is $\mathcal{O}(n^3)$, whereas the computational complexity of convolutional layers is only $\mathcal{O}(n^2)$. However, the low-rank prior enables the network to better utilize the redundancy in the data, thus facilitating improved image reconstruction \cite{zhang2024t2lr}. Visualization results under radial-16 are shown in Fig.\ref{fig:radial}. We display the reconstructed images of a particular data from the test set in the figure, along with the reconstruction results of a specific frame and x-t slice. It can be observed that our JotlasNet can reconstruct clearer textures while preserving details better. From the error maps, our network achieves better reconstruction performance in regions with more complex tissue and more intense motion.

% Tab.generated by Excel2LaTeX from sheet 'Sheet1'
\begin{table}[htbp]
	\centering
	\caption{Reconstruction Performance on OCMR Dataset (single-coil). The metrics are reported by the mean PSNR/SSIM on the test set. The inference time is in seconds. }
	\resizebox{\linewidth}{!}{\begin{tabular}{cccccc}
	  \toprule
			& radial-16 & radial-30 & vds-8 & vds-10 & TIME \\
	  \midrule
	  JotlasNet & \textbf{41.708/0.982} & \textbf{43.231/0.987} & \textbf{41.671/0.981} & \textbf{34.979/0.949} & 0.92 \\
	  L+S-Net & {41.197/0.979} & {43.013/0.986} & -     & -     & 0.32 \\
	  ISTA-Net & 40.816/0.978 & 42.962/0.986 & {41.107/0.979} & {34.793/0.948} & 0.18 \\
	  DCCNN & 40.671/0.977 & 42.897/0.986 & 40.659/0.976 & 34.599/0.947 & 0.15 \\
	  SLR-Net & 38.635/0.964 & 40.212/0.971 & 39.056/0.968 & 33.880/0.934 & 0.36 \\
	  \bottomrule
	  \end{tabular}}%
	\label{tab:sigcoil}%
  \end{table}%
  
\subsection{Multi-coil Results}
\label{sec:multicoil}
\begin{figure*}
	\centering
	\includegraphics[width=1\textwidth]{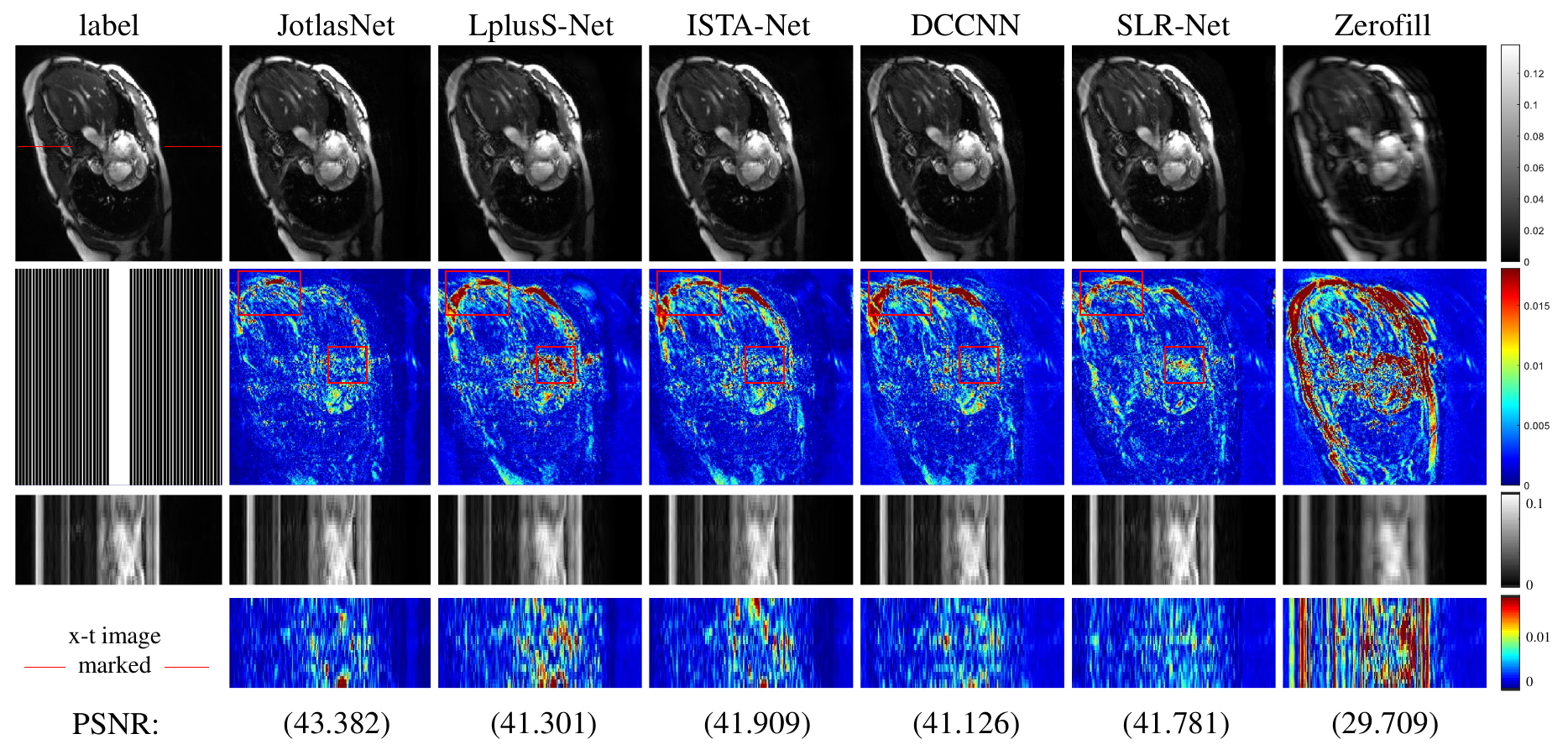}
	\caption{Reconstruction results on the CMRxRecon dataset (multi-coil) using 4X equispaced sampling. The first row shows the reconstructed images at a specific time frame, and the second row shows the corresponding error maps, except for the first image depicted the sampling mask. The third and fourth rows show the x-t images and the corresponding error maps, respectively. The position of x-t images is marked with red lines on the label image. The PSNR values respected to this test image are also listed.}
	\label{fig:equispaced}
\end{figure*}

% 我们在表4和表5中列出了各种重建方法在OCMR和CMRxRecon数据集(多线圈)上的重建结果。从定量指标中可以看到，我们的方法能够取得超过次好方法之多1dB以上的PSNR的提高。在图三中，我们展示了OCMR数据集中8倍VISTA采样模板下的重建结果。在图4中，我们展示了CMRxRecon数据集中4倍equispaced采样模板下的重建结果。从结果中可知，我们的JotlasNet能够取得得到误差更低的重建图像，同时能够更好的保留细节。我们在可视化图像中以红色方框标出了误差图中对比明显的区域。在心脏区域，我们的网络能够更好的还原心脏的形状和纹理，同时也能够更好的保留心脏的边缘。另外，我们在超高加速倍数的情况下（24倍VISTA欠采样）依然能够有效的重建图像并取得超过SOTA方法的重建效果。
In Tables \ref{tab:multicoil} and \ref{tab:cmrxrecon}, we present the quantitative results of various methods on the OCMR and CMRxRecon datasets (multi-coil). From the quantitative metrics, it can be observed that our method achieves a significant improvement of over 1dB in PSNR compared to the second-best method. 
% In Fig.\ref{fig:vista}, we show the visualized results under 8-fold VISTA sampling patterns on the OCMR dataset. 
In Fig.\ref{fig:equispaced}, we display the reconstruction results under 4X equispaced sampling patterns on the CMRxRecon dataset. Our JotlasNet can produce reconstructed images with lower errors while better preserving details. We highlight regions with noticeable differences in the error maps using red boxes in the visualized images. In the cardiac region, our network excels in restoring the shape and texture of the heart and preserving its edges. Additionally, even under ultra-high acceleration factors (e.g., 24-fold VISTA undersampling), our method effectively reconstructs images and outperforms SOTA methods.

% Tab.generated by Excel2LaTeX from sheet 'Sheet1'
\begin{table}[htbp]
	\centering
	\caption{Reconstruction Performance on OCMR Dataset (multi-coil). The metrics are reported by the mean PSNR/SSIM on the test set. }
	  \begin{tabular}{cccc}
	  \toprule
			& VISTA-8 & VISTA-12 & VISTA-24 \\
	  \midrule
	  JotlasNet & \textbf{37.508/0.971} & \textbf{36.262/0.957} & \textbf{34.030/0.948} \\
	  L+S-Net & 35.504/0.954 & 35.134/0.953 & {33.330/0.942} \\
	  ISTA-Net & {36.576/0.964} & {35.956/0.962} & 33.288/0.947 \\
	  DCCNN & 35.095/0.928 & 34.227/0.923 & 31.726/0.903 \\
	  SLR-Net & 34.212/0.953 & 32.151/0.936 & 28.512/0.891 \\
	  \bottomrule
	  \end{tabular}%
	\label{tab:multicoil}%
  \end{table}%

% Tab.generated by Excel2LaTeX from sheet 'Sheet1'
\begin{table}[htbp]
	\centering
	\caption{Reconstruction Performance on CMRxRecon Dataset (multi-coil). The metrics are reported by the mean PSNR/SSIM on the test set.}
	  \begin{tabular}{cccc}
	  \toprule
			& 4X    & 8X    & 10X \\
	  \midrule
	  JotlasNet & \textbf{38.228/0.961} & \textbf{35.631/0.951} & \textbf{34.960/0.948} \\
	  L+S-Net & 37.623/0.945 & 35.332/0.933 & 34.663/0.930 \\
	  ISTA-Net & {37.832/0.953} & {35.444/0.942} & {34.792/0.939} \\
	  DCCNN & 36.744/0.927 & 34.757/0.916 & 34.199/0.913 \\
	  SLR-Net & 37.464/0.942 & 34.709/0.926 & 34.019/0.922 \\
	  \bottomrule
	  \end{tabular}%
	\label{tab:cmrxrecon}%
  \end{table}%  

% \begin{figure*}
% 	\centering
% 	\includegraphics[width=1\textwidth]{vista.pdf}
% 	\caption{Reconstruction results on the OCMR dataset (multi-coil) using 8-fold vista sampling. The first row shows the reconstructed images at a specific time frame, and the second row shows the corresponding error maps, except for the first image depicted the sampling mask. The third and fourth rows show the x-t images and the corresponding error maps, respectively. The position of x-t images is marked with red lines on the label image. The PSNR values respected to this test image are also listed.}
% 	\label{fig:vista}
% \end{figure*}

\begin{figure}
	\centering
	\includegraphics[width=0.45\textwidth]{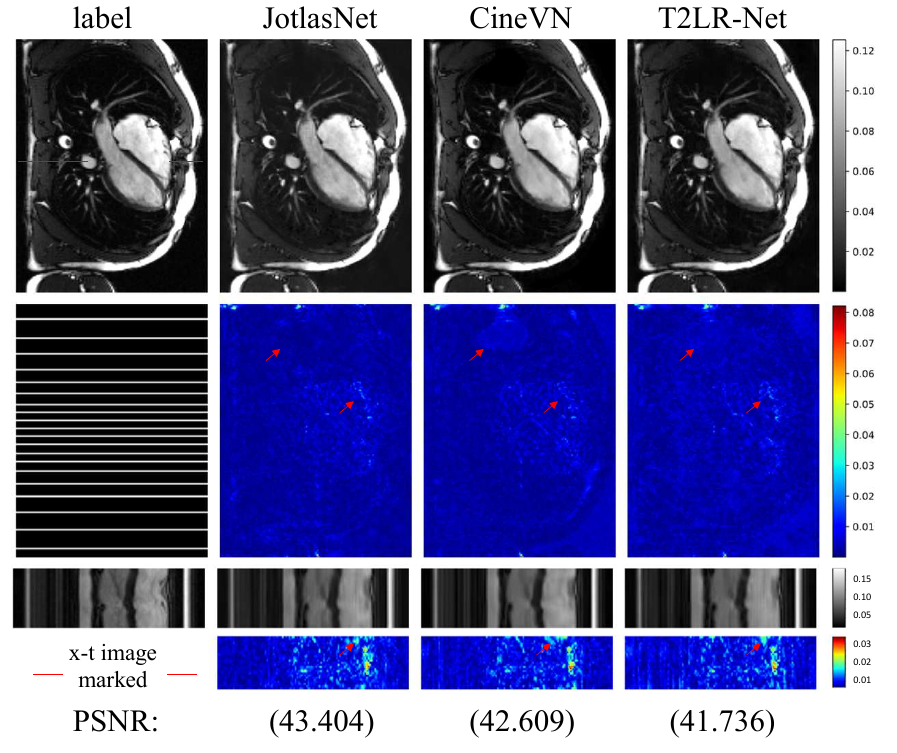}
	\caption{Reconstruction results of JotlasNet, CineVN, and T2LR-Net on the OCMR dataset (multi-coil) using 8-fold VISTA sampling.}
	\label{fig:cinevn}
\end{figure}

We further compare our JotlasNet with CineVN model (MRM 2025) \cite{vornehm2025cinevn}, which extends the end-to-end variation network \cite{sriram2020end} into spatiotemporal dynamic MRI. Another SOTA method, T2LR-Net (CIBM 2024) \cite{zhang2024t2lr}, is also compared. CineVN holds a total of 2.7M parameters (about 4 times more than JotlasNet), and is trained on the OCMR dataset with four NVIDIA A100 GPUs in parallel. Since the data splitting is different with ours, We selected 10 data samples from the intersection of our test set and the CineVN test set for evaluation. The quantitative results are listed in Tab.\ref{tab:cinevn}. The visualization results are shown in Fig.\ref{fig:cinevn}.
It is observed that our JotlasNet achieves comparable performance versus CineVN but with much fewer parameters and computation resources, which may attribute to the utilization of the tensor low-rank and sparse priors in JotlasNet while CineVN only employs the UNet to implicitly learn the spatiotemporal prior information.

\begin{table}[htbp]
	\centering
	\caption{Reconstruction Performance of JotlasNet, CineVN, and T2LR-Net. The metrics are reported by mean ± standard derivation.}
	\resizebox{\linewidth}{!}{\begin{tabular}{ccccc}
	  \toprule
			& \multicolumn{2}{c}{VISTA-8} & \multicolumn{2}{c}{VISTA-12} \\
			& PSNR  & SSIM  & PSNR  & SSIM \\
	  \midrule
	  JotlasNet & 40.791±3.733 & 0.981±0.014 & 40.047±3.161 & 0.973±0.014 \\
	  CineVN (MRM 2025) & 40.975±4.319 & 0.981±0.006 & 40.349±4.186 & 0.979±0.006 \\
	  T2LR-Net (CIBM 2024) & 40.404±4.602 & 0.978±0.012 & 39.856±4.299 & 0.963±0.018 \\
	  \bottomrule
	  \end{tabular}}%
	\label{tab:cinevn}%
  \end{table}%

\subsection{Abalation Study}
\label{sec:ablation}
% 我们在radial-16采样模板情况下利用OCMR单线圈数据进行消融实验，如表6所示。第一行表示我们的JotlasNet，第二行是利用SLR提出的算法，如公式(1)，来求解模型(3)并展开成网络，第三行和第四行是去掉低秩先验的网络，区别在于，第三行使用了我们所提出的AST算子，而第四行使用传统的只有一个阈值的ST算子。从第一行和第二行的对比中可以发现，我们所提出的算法1所展开的网络能够取得更好的重建效果，并且，算法1具有同单先验ISTA非常相似的简单又高效的结构。从第一行和第三行的对比中可以发现，双先验网络能够取得比单一先验更优的结果。从最后两行的对比中可以发现，我们所提出的AST算子能够切实提高网络利用稀疏的能力，从而取得更好的重建效果。
Under the radial-16 case, we conducted ablation experiments using OCMR single-coil data, as shown in Tab.\ref{tab:abalation}. Model-1 represents our JotlasNet; model-2 utilizes the algorithm proposed by SLR-Net, as described in \eqref{eq:opt_pgdadmm}, to solve our proposed tensor low-rank and sparse model and unfold it into a network. Model-3 is the network without the sparse prior. Model-4 and 5 represent networks without the low-rank prior, which are unfolded from the ISTA algorithm as \eqref{eq:ista}. The distinction is that model-4 employs the AST operator proposed by us, while model-5 uses the traditional ST operator with only one threshold. From the comparison between model-1 and 2, it can be observed that the network unfolded by our proposed Alg.\ref{algo_cps} achieves better reconstruction results since Alg.\ref{algo_cps} exhibits a simple yet efficient structure similar to the single-prior ISTA. Comparing model-1 with 3 and 4, it can be seen that the dual-prior network achieves superior results compared to the single-prior network. From the comparison between model-4 and 5, it can be concluded that our proposed AST operator effectively enhances the network's ability to exploit sparsity, thereby achieving better reconstruction results.

% 如3.2节所提到的，我们对学习低秩和稀疏变换域的前后两个CNN不做任何约束。我们以model 3为基准来分析CNN约束对网络的影响，原因是model 3只使用稀疏先验因此只涉及一组CNN学习的变换，即公式16中的D1和D2，能够控制变量因此更清晰的分析CNN约束对网络的影响。
As mentioned in Section \ref{sec:net}, the CNNs learning the low-rank or sparse transformations are subjected to no constraints in our implementation. We use model-4 as the baseline to analyze the impact of CNN constraints on the network. 
% The reason is that model 3 only employs a sparse prior, thus involving only one set of CNN-learned transforms, namely $\tilde{\mathsf{D}}_n$ and $\mathsf{D}_n$ in \eqref{eq:net}. This allows us to control variables and make the analysis more clearly.
% 我们在model 3的MSE loss的基础上增加了CNN可逆约束，就像ISTA-Net和SLR-Net那样，即新的loss为：
Specifically, on top of the MSE loss in model-4, we incorporate the inverse constraint on the CNN-learned sparse transforms, akin to ISTA-Net \cite{ref_ISTANET} and SLR-Net \cite{ref_slrnet},
\begin{equation}
	\label{loss_func2}
	\begin{aligned}
		Loss = \sum_{(\mathcal{X}_{GT}, \mathbf{b}) \in \Omega}&\Vert \mathcal{X}_{GT} - f_{net}(\mathbf{b}|\theta) \Vert_F^2 \\+ \zeta &\sum_{n=1}^{N}\Vert \tilde{\mathsf{D}}_n \circ \mathsf{D}_n(\mathcal{X}_{n-1}) - \mathcal{X}_{n-1} \Vert_F^2,
	\end{aligned}
  \end{equation}
where $\zeta$ is the hyperparameter controlling the strength of the inverse constraint. We set $\zeta$ to 0.001, 0.01, and 0.1 to investigate the impact of the CNN constraint on the network. The results are shown in model-6 to 8. Compared with model-4 where $\zeta=0$, it can be observed that the network's performance deteriorates as the CNN constraint strengthens. This indicates that inverse constraint forces CNNs to solely learn explicit sparse characteristics, without further integrating their implicit feature extraction capabilities.

\begin{figure}
	\centering
	\includegraphics[width=0.4\textwidth]{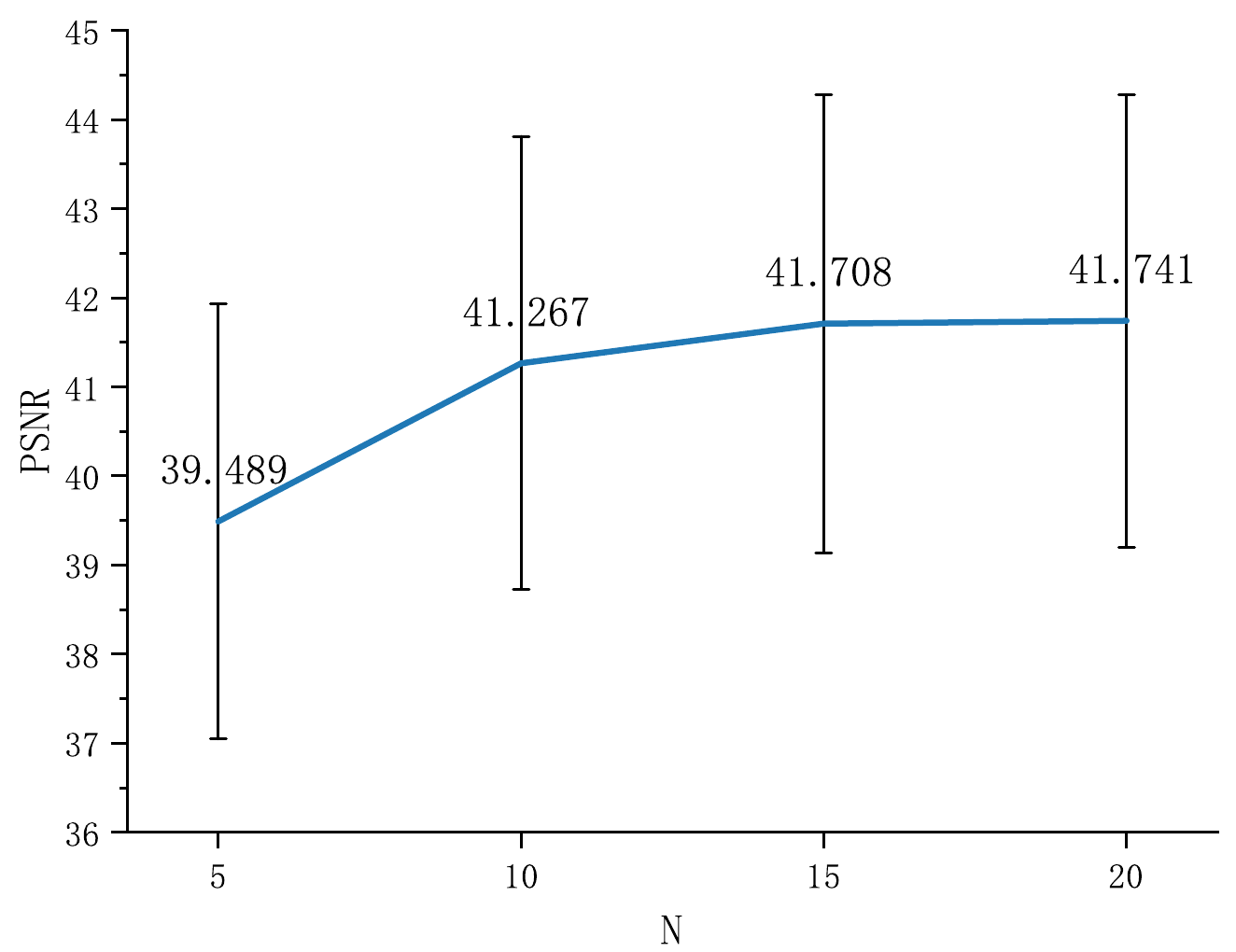}
	\caption{The PSNR of JotlasNet under different numbers of iterations (N). The blue line shows the mean PSNRs while the error bars represents the standard deviations.}
	\label{fig:iters}
\end{figure}

Additionally, we also investigated the optimal number of iterations of JotlasNet under the radial-16 pattern in single coil scenario, as the results are shown in Fig.\ref{fig:iters}. It can be observed that the PSNR increase slowly when the number of iterations exceeds 15. Therefore, in order to balance the reconstruction performance and computational efficiency, we set the number of iterations to 15 in our experiments.

% Tab.generated by Excel2LaTeX from sheet 'Sheet1'
\begin{table*}[htbp]
	\centering
	\caption{Abalation study. The metrics are reported as mean ± standard deviation.}
	  \begin{tabular}{ccccccc}
	  \toprule
	  model num & unrolled from & low rank & sparse & CNN constraint & PSNR  & SSIM \\
	  \midrule
	  1(ours) & Alg. 1 & \checkmark     & AST   & 0     & 41.708 ± 2.569 & 0.982 ± 0.007 \\
	  2     & SLR-Net Alg. & \checkmark     & AST   & 0     & 41.629 ± 2.576 & 0.981 ± 0.007 \\
	  3     & ISTA  & \checkmark     & $\times$   & 0     & 40.788 ± 2.520 & 0.978 ± 0.008 \\
	  4     & ISTA  & $\times$     & AST   & 0     & 41.617 ± 2.582 & 0.981 ± 0.008 \\
	  5     & ISTA  & $\times$     & ST    & 0     & 41.428 ± 2.592 & 0.980 ± 0.008 \\
	  6     & ISTA  & $\times$     & AST   & 0.001 & 41.588 ± 2.618 & 0.981 ± 0.008 \\
	  7     & ISTA  & $\times$     & AST   & 0.01  & 40.950 ± 2.528 & 0.977 ± 0.008 \\
	  8     & ISTA  & $\times$     & AST   & 0.1   & 36.997 ± 2.253 & 0.947 ± 0.014 \\
	  \bottomrule
	  \end{tabular}%
	\label{tab:abalation}%
  \end{table*}%

\section{Discussion}
\subsection{Interpretive Insight of LR and S layers}
\begin{figure*}
	\centering
	\includegraphics[width=0.9\textwidth]{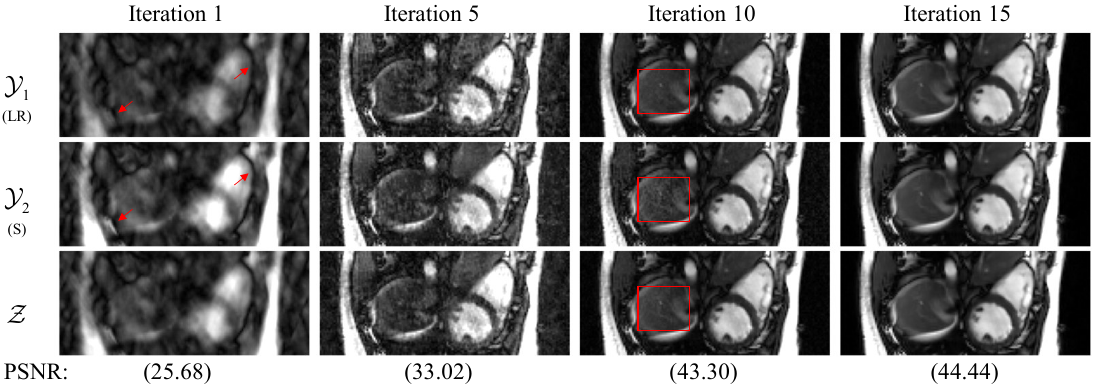}
	\caption{The intermediate results from different iterations of JotlasNet. The symbols are the same as Alg.\ref{algo_cps}. The LR and S layers as well as their combination are shown along the vertical direction while the horizontal axis represents the different iterations. The corresponding PSNRs are also listed below the images. \label{fig:abla}}
\end{figure*}
We show the intermediate results from different iterations of JotlasNet under radial-16 pattern in single coil scenario in Fig.\ref{fig:abla}. In the early iterations (1-5), the LR layer focuses on the background while the S layer captures the edges, as indicated by the red arrows. In the middle iterations (6-10), the LR layer tends to smooth the textures due to its ability to capture correlation information, while the S layer adds the image details inside the smooth regions, as indicated by the red boxes. In the later iterations (11-15), the LR along with the S layer jointly remove the noise and refine the textures, as the PSNR increases slowly.

\subsection{Rethinking the Structure of Neural Networks}
\begin{figure}
	\centering
	\includegraphics[width=0.7\linewidth]{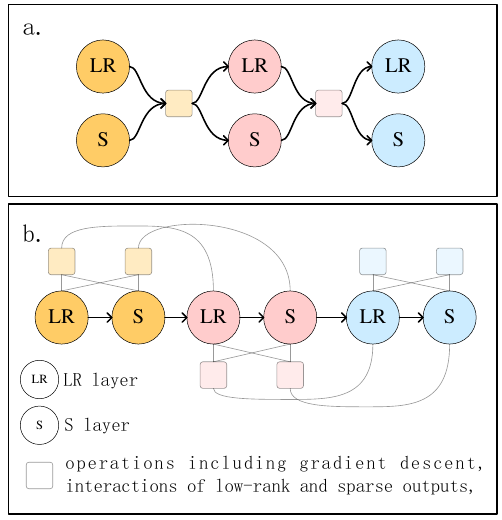}
	\caption{Simplified diagrams of the networks unfolded from equations \eqref{eq:net} and \eqref{eq:opt_lpluss}. (a) The network unfolded from equation \eqref{eq:net}, representing the simplified diagram of our proposed network. (b) The network unfolded from equation \eqref{eq:opt_lpluss}.}
	\label{fig:structure}
\end{figure}

% 我们将联合低秩和稀疏先验的composite splitting algorithm和利用低秩加稀疏的ISTA算法以及两种算法展开的深度展开网络进行对比分析，以重新思考神经网络的架构。前者，我们已经在前面进行了详细的叙述，而后者其利用以下模型并可以利用ISTA直接求解得到：
We compare the composite splitting algorithm that incorporates joint low-rank and sparsity priors with the ISTA algorithm that utilizes low-rank plus sparse priors. We analyze the DUNs derived from these two algorithms to reconsider neural network architecture. The former has been elaborately described earlier, while the latter utilizes the following model and can be directly solved by ISTA \cite{huang2021deep}:
\begin{equation}
	\min_{\mathcal{X}} \frac12 \| \mathsf{A}(\mathcal{L} + \mathcal{S}) - \mathbf{b} \|_2^2 + \lambda_1 \mathsf{R}(\mathcal{L}) + \lambda_2 \mathsf{S}(\mathcal{S}),
\end{equation}
\begin{equation}
	\label{eq:opt_lpluss}
	\begin{cases}
		\bar{\mathcal{L}} &= \mathcal{L}^{n}-{\mu} \mathsf{A}^H\left[\mathsf{A}(\mathcal{L}^{n}+\mathcal{S}^{n})-\mathbf{b}\right] \\
		\bar{\mathcal{S}} &= \mathcal{S}^{n}-{\mu} \mathsf{A}^H\left[\mathsf{A}(\mathcal{L}^{n}+\mathcal{S}^{n})-\mathbf{b}\right] \\
		\mathcal{L}^{n+1} &= \operatorname{prox}_{\mu\lambda_1 \mathsf{R}}(\bar{\mathcal{L}}-\mathcal{S}^n)\\
		\mathcal{S}^{n+1} &= \operatorname{prox}_{\mu\lambda_2 \mathsf{S}}(\bar{\mathcal{S}}-\mathcal{L}^{n+1})\\
	\end{cases},
\end{equation}
% 我们把关注的重点放在式（16）和（21）。两式展开为网络的简化示意图如图5所示，其中，图5a展示了（16）展开的结果，即我们所提出的网络的简化图，图5b展示了（21）展开的结果。对低秩约束和稀疏约束的映射步在网络中用LR和S layers来表示，对其他的涉及梯度下降，低秩和稀疏输出的交互用半透明的方块表示，迭代之间的联系用半透明的线表示。我们将关注重点放在含有神经网络的LR和S layers上，我们发现，对于联合利用低秩和稀疏先验的展开自（16）的网络，LR和S layers呈现并联的形式，而对于低秩和稀疏分离的展开自（21）的网络，LR和S layers呈现串联的形式。这一发现可能对神经网络的结构提供了部分可解释性，同时对深度展开网络的结构设计提供了一定的启示。
The simplified diagrams of the networks unfolded from equations \eqref{eq:net} and \eqref{eq:opt_lpluss} are illustrated in Fig.\ref{fig:structure}. Fig.\ref{fig:structure}a depicts the structure of unfolding equation \eqref{eq:net}, representing the simplified diagram of our proposed network, while Fig.\ref{fig:structure}b illustrates the structure of unrolling equation \eqref{eq:opt_lpluss}. The projection operations associated with low-rank and sparse constraints are represented by LR and S layers, respectively. Operations involving gradient descent and interactions between low-rank and sparse outputs are depicted by translucent squares, and interactions between iterations are represented by translucent lines. It is observed that for the network unfolded from equation \eqref{eq:net}, which exploits the joint utilization of low-rank and sparse priors, LR and S layers appear in a parallel configuration. Conversely, for the network unfolded from equation \eqref{eq:opt_lpluss}, which separates low-rank and sparse parts, LR and S layers appear in a serial configuration. This observation may provide partial interpretability to the structure of neural networks and offer insights for the design of DUNs.

\subsection{Limitations and Future Work} 
% 我们所提出的JotlasNet对低秩和稀疏先验进行联合利用。低秩先验的利用往往需要进行SVD的计算，而就像4.2节所提到的，SVD的梯度具有数值不稳定性，特别是当有两个重复奇异值出现时，会出现NaN梯度，导致训练无法进行。由于需要进行上万步的训练，这种风险出现的概率较大。因此，我们将在未来的工作中研究SVD的梯度计算和致力于开发可微分的SVD来解决这个问题。另外，SVD的计算复杂度较高，会增加对网络推理时间。因此，我们也将研究无需SVD的低秩先验的利用方法，并结合到深度展开网络当中。
Our proposed JotlasNet leverages jointly low-rank and sparse prior. Utilizing the low-rank prior often involves computing SVD. However, as mentioned in Section \ref{sec:sigcoil}, the gradients of SVD exhibit numerical instability, particularly when two repeated singular values occur, leading to NaN gradients and hindering the training process \cite{wang2021robust}. Given the necessity for thousands of training steps, this risk is considerable. Hence, we plan to investigate gradient computation techniques for SVD and focus on developing differentiable SVD to address this issue in future work. Additionally, the computational complexity of SVD is high, which increases the inference time of the network. Therefore, we will also explore methods for utilizing low-rank prior without relying on SVD \cite{liang2007spatiotemporal} and integrate them into DUNs.

% 另外，在本论文中，我们选取了低秩和稀疏两种在过去的MRI重建中被广泛使用并证实有效的先验来构建JotlasNet。实际上，在composite splitting algorithm的框架下，我们可以将LR和S layers替换为其他的先验，例如TV先验，时间域傅里叶变换稀疏先验等传统的先验，当然，也可以结合当前对于神经网络的前沿研究，例如引入transformer提取并利用整体特征先验，而用CNN来提取局部特征先验等。这些将是我们未来的研究方向。
Furthermore, in this paper, we chose low-rank and sparse priors, two widely used and proven effective priors in past MRI reconstruction, to construct JotlasNet. In fact, within the framework of the composite splitting algorithm, we can replace LR and S layers with other priors, such as total variation prior \cite{lingala2011accelerated}, sparse prior in temporal Fourier domain \cite{jung2009k}, and other traditional priors. Moreover, we can integrate cutting-edge research in neural networks, such as introducing transformers 
%\cite{dosovitskiy2020image} 
to extract and utilize global feature priors, while using CNNs to extract local feature priors. These avenues will be the focus of our future research.

Additionally, many works have incorporated a coil sensitivity map estimation (SME) module to refine the CSMs at each iteration, which further improves the reconstruction performance \cite{sriram2020end,fabian2022humus}, especially when autocalibration data is limited. We will explore the integration of the SME module into JotlasNet to enhance the reconstruction performance in the future work.

\section{Conclusion}
% 在本文中，我们提出了novel JotlasNet。这是联合可学习变换域张量低秩和稀疏先验的首次尝试。张量低秩先验基于t-SVD分解模型构建，稀疏先验基于变换域l1范数实现。变换域由CNNs来依照监督机制基于数据集学习。我们基于composite splitting algorithm设计了求解模型的迭代优化算法，并进一步使得展开此算法后的展开网络具有一个简单但高效的并联结构。另外，我们还提出了一种全新的AST算子，用来替代进行稀疏约束的ST算子。AST能够利用自注意力机制为每个通道的数据单独学习一个阈值，提高了网络对稀疏先验的利用灵活性。我们在OCMR和CMRxRecon数据集上进行了大量实验，结果表明，我们的JotlasNet在多线圈和单线圈的MRI重建任务中取得了最好的重建效果。我们还进行了消融实验和结构分析，结果表明，我们所提出的网络结构能够取得更好的重建效果，并且对神经网络的结构设计提供了一定的启示。
In this paper, we introduced a novel JotlasNet, representing the first attempt to jointly utilize transformed tensor low-rank and sparse priors. The low-rank and sparse domains are learned by CNNs. High-dimensional data correlations are exploited through the tensor nuclear norm. A novel AST operator was proposed to replace the conventional ST operator for sparse constraint. AST utilizes self-attention mechanism to learn individual thresholds for each channel, enhancing the flexibility of the network in utilizing sparse priors.
Additionally, we designed an iterative algorithm based on the composite splitting algorithm to solve the model, further unrolling it into the proposed JotlasNet. The network exhibits a simple yet efficient parallel structure. Extensive experiments on the OCMR and CMRxRecon datasets demonstrated that our JotlasNet achieved superior reconstruction performance in both multi-coil and single-coil MRI reconstruction tasks. We posit that the framework we proposed carries profound implications for various models incorporating joint priors, extending beyond the interaction of low-rank and sparse priors and transcending the realm of dynamic MRI reconstruction applications.

%%%%%%%%%%%%%%%%%%%%%%%%%%%%%%%%%%%%%%%%%%%%%%%%%%%%%%%%%%%%%%%%%%%%%%%%%
\section*{Acknowledgments}
This work is supported by the National Natural Science Foundation of China [grant number 62371167]; the Natural Science Foundation of Heilongjiang [grant number YQ2021F005].

% \section*{References}
\bibliographystyle{model1-num-names.bst}
\bibliography{refs}  %%% Uncomment this line and comment out the ``thebibliography'' section below to use the external .bib file (using bibtex) .

\begin{thebibliography}{49}
\expandafter\ifx\csname natexlab\endcsname\relax\def\natexlab#1{#1}\fi
\providecommand{\url}[1]{\texttt{#1}}
\providecommand{\href}[2]{#2}
\providecommand{\path}[1]{#1}
\providecommand{\DOIprefix}{doi:}
\providecommand{\ArXivprefix}{arXiv:}
\providecommand{\URLprefix}{URL: }
\providecommand{\Pubmedprefix}{pmid:}
\providecommand{\doi}[1]{\href{http://dx.doi.org/#1}{\path{#1}}}
\providecommand{\Pubmed}[1]{\href{pmid:#1}{\path{#1}}}
\providecommand{\bibinfo}[2]{#2}
\ifx\xfnm\relax \def\xfnm[#1]{\unskip,\space#1}\fi
%Type = Article
\bibitem[{Muckley et~al.(2021)Muckley, Riemenschneider
  et~al.}]{muckley2021results}
\bibinfo{author}{M.~J. Muckley}, \bibinfo{author}{B.~Riemenschneider}, et~al.,
\newblock \bibinfo{title}{Results of the 2020 {fastMRI} challenge for machine
  learning {MR} image reconstruction},
\newblock \bibinfo{journal}{IEEE Trans Med Imaging} \bibinfo{volume}{40}
  (\bibinfo{year}{2021}) \bibinfo{pages}{2306--2317}.
%Type = Article
\bibitem[{Lustig et~al.(2008)Lustig, Donoho, Santos, and
  Pauly}]{lustig2008compressed}
\bibinfo{author}{M.~Lustig}, \bibinfo{author}{D.~L. Donoho},
  \bibinfo{author}{J.~M. Santos}, \bibinfo{author}{J.~M. Pauly},
\newblock \bibinfo{title}{Compressed sensing {MRI}},
\newblock \bibinfo{journal}{IEEE signal processing magazine}
  \bibinfo{volume}{25} (\bibinfo{year}{2008}) \bibinfo{pages}{72--82}.
%Type = Article
\bibitem[{Lustig et~al.(2007)Lustig, Donoho, and Pauly}]{lustig2007sparse}
\bibinfo{author}{M.~Lustig}, \bibinfo{author}{D.~Donoho},
  \bibinfo{author}{J.~M. Pauly},
\newblock \bibinfo{title}{Sparse {MRI}: The application of compressed sensing
  for rapid {MR} imaging},
\newblock \bibinfo{journal}{Magnetic Resonance in Medicine}
  \bibinfo{volume}{58} (\bibinfo{year}{2007}) \bibinfo{pages}{1182--1195}.
%Type = Article
\bibitem[{Boyd et~al.(2011)Boyd, Parikh, Chu, Peleato, Eckstein
  et~al.}]{boyd2011distributed}
\bibinfo{author}{S.~Boyd}, \bibinfo{author}{N.~Parikh},
  \bibinfo{author}{E.~Chu}, \bibinfo{author}{B.~Peleato},
  \bibinfo{author}{J.~Eckstein}, et~al.,
\newblock \bibinfo{title}{Distributed optimization and statistical learning via
  the alternating direction method of multipliers},
\newblock \bibinfo{journal}{Foundations and Trends{\textregistered} in Machine
  learning} \bibinfo{volume}{3} (\bibinfo{year}{2011}) \bibinfo{pages}{1--122}.
%Type = Article
\bibitem[{Afonso et~al.(2010)Afonso, Bioucas-Dias, and
  Figueiredo}]{afonso2010admm}
\bibinfo{author}{M.~V. Afonso}, \bibinfo{author}{J.~M. Bioucas-Dias},
  \bibinfo{author}{M.~A. Figueiredo},
\newblock \bibinfo{title}{Fast image recovery using variable splitting and
  constrained optimization},
\newblock \bibinfo{journal}{IEEE Trans Image Processing} \bibinfo{volume}{19}
  (\bibinfo{year}{2010}) \bibinfo{pages}{2345--2356}.
%Type = Article
\bibitem[{Beck and Teboulle(2009)}]{beck2009fast}
\bibinfo{author}{A.~Beck}, \bibinfo{author}{M.~Teboulle},
\newblock \bibinfo{title}{A fast iterative shrinkage-thresholding algorithm for
  linear inverse problems},
\newblock \bibinfo{journal}{SIAM journal on imaging sciences}
  \bibinfo{volume}{2} (\bibinfo{year}{2009}) \bibinfo{pages}{183--202}.
%Type = Article
\bibitem[{Huang et~al.(2011)Huang, Zhang, and Metaxas}]{huang2011efficient}
\bibinfo{author}{J.~Huang}, \bibinfo{author}{S.~Zhang},
  \bibinfo{author}{D.~Metaxas},
\newblock \bibinfo{title}{Efficient {MR} image reconstruction for compressed
  {MR} imaging},
\newblock \bibinfo{journal}{Medical Image Analysis} \bibinfo{volume}{15}
  (\bibinfo{year}{2011}) \bibinfo{pages}{670--679}.
%Type = Article
\bibitem[{Jung et~al.(2009)Jung, Sung, Nayak, Kim, and Ye}]{jung2009k}
\bibinfo{author}{H.~Jung}, \bibinfo{author}{K.~Sung}, \bibinfo{author}{K.~S.
  Nayak}, \bibinfo{author}{E.~Y. Kim}, \bibinfo{author}{J.~C. Ye},
\newblock \bibinfo{title}{{k-t FOCUSS}: a general compressed sensing framework
  for high resolution dynamic {MRI}},
\newblock \bibinfo{journal}{Magnetic Resonance in Medicine}
  \bibinfo{volume}{61} (\bibinfo{year}{2009}) \bibinfo{pages}{103--116}.
%Type = Article
\bibitem[{Ravishankar and Bresler(2010)}]{ravishankar2010mr}
\bibinfo{author}{S.~Ravishankar}, \bibinfo{author}{Y.~Bresler},
\newblock \bibinfo{title}{{MR} image reconstruction from highly undersampled
  k-space data by dictionary learning},
\newblock \bibinfo{journal}{IEEE transactions on medical imaging}
  \bibinfo{volume}{30} (\bibinfo{year}{2010}) \bibinfo{pages}{1028--1041}.
%Type = Inproceedings
\bibitem[{Liang(2007)}]{liang2007spatiotemporal}
\bibinfo{author}{Z.-P. Liang},
\newblock \bibinfo{title}{Spatiotemporal imagingwith partially separable
  functions},
\newblock in: \bibinfo{booktitle}{2007 4th IEEE international symposium on
  biomedical imaging: from nano to macro}, \bibinfo{organization}{IEEE},
  \bibinfo{year}{2007}, pp. \bibinfo{pages}{988--991}.
%Type = Article
\bibitem[{He et~al.(2016)He, Liu, Christodoulou, Ma, Lam, and
  Liang}]{he2016accelerated}
\bibinfo{author}{J.~He}, \bibinfo{author}{Q.~Liu}, \bibinfo{author}{A.~G.
  Christodoulou}, \bibinfo{author}{C.~Ma}, \bibinfo{author}{F.~Lam},
  \bibinfo{author}{Z.-P. Liang},
\newblock \bibinfo{title}{Accelerated high-dimensional {MR} imaging with sparse
  sampling using low-rank tensors},
\newblock \bibinfo{journal}{IEEE Trans Med Imaging} \bibinfo{volume}{35}
  (\bibinfo{year}{2016}) \bibinfo{pages}{2119--2129}.
%Type = Article
\bibitem[{Otazo et~al.(2015)Otazo, Candes, and Sodickson}]{otazo2015low}
\bibinfo{author}{R.~Otazo}, \bibinfo{author}{E.~Candes}, \bibinfo{author}{D.~K.
  Sodickson},
\newblock \bibinfo{title}{Low-rank plus sparse matrix decomposition for
  accelerated dynamic {MRI} with separation of background and dynamic
  components},
\newblock \bibinfo{journal}{Magnetic resonance in medicine}
  \bibinfo{volume}{73} (\bibinfo{year}{2015}) \bibinfo{pages}{1125--1136}.
%Type = Article
\bibitem[{Lingala et~al.(2011)Lingala, Hu, DiBella, and
  Jacob}]{lingala2011accelerated}
\bibinfo{author}{S.~G. Lingala}, \bibinfo{author}{Y.~Hu},
  \bibinfo{author}{E.~DiBella}, \bibinfo{author}{M.~Jacob},
\newblock \bibinfo{title}{Accelerated dynamic {MRI} exploiting sparsity and
  low-rank structure: {k-t SLR}},
\newblock \bibinfo{journal}{IEEE Trans Med Imaging} \bibinfo{volume}{30}
  (\bibinfo{year}{2011}) \bibinfo{pages}{1042--1054}.
%Type = Inproceedings
\bibitem[{Gregor and LeCun(2010)}]{LeCun2010unroll}
\bibinfo{author}{K.~Gregor}, \bibinfo{author}{Y.~LeCun},
\newblock \bibinfo{title}{Learning fast approximations of sparse coding},
\newblock in: \bibinfo{booktitle}{Proceedings of the 27th international
  conference on international conference on machine learning},
  \bibinfo{year}{2010}, pp. \bibinfo{pages}{399--406}.
%Type = Article
\bibitem[{Ke et~al.(2021)Ke, Huang, Cui, Cheng, Jia, Wang, Liu, Zheng, Ying,
  Zhu et~al.}]{ref_slrnet}
\bibinfo{author}{Z.~Ke}, \bibinfo{author}{W.~Huang}, \bibinfo{author}{Z.-X.
  Cui}, \bibinfo{author}{J.~Cheng}, \bibinfo{author}{S.~Jia},
  \bibinfo{author}{H.~Wang}, \bibinfo{author}{X.~Liu},
  \bibinfo{author}{H.~Zheng}, \bibinfo{author}{L.~Ying},
  \bibinfo{author}{Y.~Zhu}, et~al.,
\newblock \bibinfo{title}{Learned low-rank priors in dynamic {MR} imaging},
\newblock \bibinfo{journal}{IEEE Trans Med Imaging} \bibinfo{volume}{40}
  (\bibinfo{year}{2021}) \bibinfo{pages}{3698--3710}.
%Type = Article
\bibitem[{Schlemper et~al.(2017)Schlemper, Caballero, Hajnal, Price, and
  Rueckert}]{ref_DCCNN}
\bibinfo{author}{J.~Schlemper}, \bibinfo{author}{J.~Caballero},
  \bibinfo{author}{J.~V. Hajnal}, \bibinfo{author}{A.~N. Price},
  \bibinfo{author}{D.~Rueckert},
\newblock \bibinfo{title}{A deep cascade of convolutional neural networks for
  dynamic {MR} image reconstruction},
\newblock \bibinfo{journal}{IEEE Trans Med Imaging} \bibinfo{volume}{37}
  (\bibinfo{year}{2017}) \bibinfo{pages}{491--503}.
%Type = Inproceedings
\bibitem[{Zhang and Ghanem(2018)}]{ref_ISTANET}
\bibinfo{author}{J.~Zhang}, \bibinfo{author}{B.~Ghanem},
\newblock \bibinfo{title}{{ISTA-Net}: Interpretable optimization-inspired deep
  network for image compressive sensing},
\newblock in: \bibinfo{booktitle}{IEEE CVPR conference}, \bibinfo{year}{2018},
  pp. \bibinfo{pages}{1828--1837}.
%Type = Article
\bibitem[{Donoho(1995)}]{donoho1995noising}
\bibinfo{author}{D.~L. Donoho},
\newblock \bibinfo{title}{De-noising by soft-thresholding},
\newblock \bibinfo{journal}{IEEE transactions on information theory}
  \bibinfo{volume}{41} (\bibinfo{year}{1995}) \bibinfo{pages}{613--627}.
%Type = Article
\bibitem[{Lu et~al.(2019)Lu, Feng, Chen, Liu, Lin, and Yan}]{lu2019tensor}
\bibinfo{author}{C.~Lu}, \bibinfo{author}{J.~Feng}, \bibinfo{author}{Y.~Chen},
  \bibinfo{author}{W.~Liu}, \bibinfo{author}{Z.~Lin}, \bibinfo{author}{S.~Yan},
\newblock \bibinfo{title}{Tensor robust principal component analysis with a new
  tensor nuclear norm},
\newblock \bibinfo{journal}{IEEE transactions on pattern analysis and machine
  intelligence} \bibinfo{volume}{42} (\bibinfo{year}{2019})
  \bibinfo{pages}{925--938}.
%Type = Article
\bibitem[{Kilmer and Martin(2011)}]{ref_tsvd}
\bibinfo{author}{M.~E. Kilmer}, \bibinfo{author}{C.~D. Martin},
\newblock \bibinfo{title}{Factorization strategies for third-order tensors},
\newblock \bibinfo{journal}{Linear Algebra and its Applications}
  \bibinfo{volume}{435} (\bibinfo{year}{2011}) \bibinfo{pages}{641--658}.
%Type = Article
\bibitem[{Zhang et~al.(2024)Zhang, Li, and Hu}]{zhang2024t2lr}
\bibinfo{author}{Y.~Zhang}, \bibinfo{author}{P.~Li}, \bibinfo{author}{Y.~Hu},
\newblock \bibinfo{title}{{T$^2$LR-Net}: An unrolling network learning
  transformed tensor low-rank prior for dynamic {MR} image reconstruction},
\newblock \bibinfo{journal}{Computers in Biology and Medicine}
  (\bibinfo{year}{2024}) \bibinfo{pages}{108034}.
%Type = Inproceedings
\bibitem[{Zhang et~al.(2020)Zhang, Liu, Wu, and Walid}]{zhang2020video}
\bibinfo{author}{Y.~Zhang}, \bibinfo{author}{X.-Y. Liu},
  \bibinfo{author}{B.~Wu}, \bibinfo{author}{A.~Walid},
\newblock \bibinfo{title}{Video synthesis via transform-based tensor neural
  network},
\newblock in: \bibinfo{booktitle}{Proceedings of the 28th ACM International
  Conference on Multimedia}, \bibinfo{year}{2020}, pp.
  \bibinfo{pages}{2454--2462}.
%Type = Article
\bibitem[{Kolda and Bader(2009)}]{kolda2009tensor}
\bibinfo{author}{T.~G. Kolda}, \bibinfo{author}{B.~W. Bader},
\newblock \bibinfo{title}{Tensor decompositions and applications},
\newblock \bibinfo{journal}{SIAM review} \bibinfo{volume}{51}
  (\bibinfo{year}{2009}) \bibinfo{pages}{455--500}.
%Type = Article
\bibitem[{Huang et~al.(2011)Huang, Zhang, Li, and Metaxas}]{huang2011composite}
\bibinfo{author}{J.~Huang}, \bibinfo{author}{S.~Zhang},
  \bibinfo{author}{H.~Li}, \bibinfo{author}{D.~Metaxas},
\newblock \bibinfo{title}{Composite splitting algorithms for convex
  optimization},
\newblock \bibinfo{journal}{Computer Vision and Image Understanding}
  \bibinfo{volume}{115} (\bibinfo{year}{2011}) \bibinfo{pages}{1610--1622}.
%Type = Inproceedings
\bibitem[{Zhang et~al.(2023)Zhang, Li, Li, and Hu}]{zhang2023dusnet}
\bibinfo{author}{Y.~Zhang}, \bibinfo{author}{X.~Li}, \bibinfo{author}{W.~Li},
  \bibinfo{author}{Y.~Hu},
\newblock \bibinfo{title}{Deep unrolling shrinkage network for dynamic {MR}
  imaging},
\newblock in: \bibinfo{booktitle}{2023 IEEE International Conference on Image
  Processing (ICIP)}, \bibinfo{year}{2023}, pp. \bibinfo{pages}{1145--1149}.
  \DOIprefix\doi{10.1109/ICIP49359.2023.10223077}.
%Type = Article
\bibitem[{Liang et~al.(2020)Liang, Cheng, Ke, and Ying}]{liang2020deep}
\bibinfo{author}{D.~Liang}, \bibinfo{author}{J.~Cheng},
  \bibinfo{author}{Z.~Ke}, \bibinfo{author}{L.~Ying},
\newblock \bibinfo{title}{Deep magnetic resonance image reconstruction: Inverse
  problems meet neural networks},
\newblock \bibinfo{journal}{IEEE Signal Processing Magazine}
  \bibinfo{volume}{37} (\bibinfo{year}{2020}) \bibinfo{pages}{141--151}.
%Type = Article
\bibitem[{Monga et~al.(2021)Monga, Li, and Eldar}]{monga2021algorithm}
\bibinfo{author}{V.~Monga}, \bibinfo{author}{Y.~Li}, \bibinfo{author}{Y.~C.
  Eldar},
\newblock \bibinfo{title}{Algorithm unrolling: Interpretable, efficient deep
  learning for signal and image processing},
\newblock \bibinfo{journal}{IEEE Signal Processing Magazine}
  \bibinfo{volume}{38} (\bibinfo{year}{2021}) \bibinfo{pages}{18--44}.
%Type = Article
\bibitem[{Aggarwal et~al.(2018)Aggarwal, Mani, and Jacob}]{aggarwal2018modl}
\bibinfo{author}{H.~K. Aggarwal}, \bibinfo{author}{M.~P. Mani},
  \bibinfo{author}{M.~Jacob},
\newblock \bibinfo{title}{{MoDL}: Model-based deep learning architecture for
  inverse problems},
\newblock \bibinfo{journal}{IEEE Trans Med Imaging} \bibinfo{volume}{38}
  (\bibinfo{year}{2018}) \bibinfo{pages}{394--405}.
%Type = Inproceedings
\bibitem[{Sriram et~al.(2020)Sriram, Zbontar, Murrell, Defazio, Zitnick,
  Yakubova, Knoll, and Johnson}]{sriram2020end}
\bibinfo{author}{A.~Sriram}, \bibinfo{author}{J.~Zbontar},
  \bibinfo{author}{T.~Murrell}, \bibinfo{author}{A.~Defazio},
  \bibinfo{author}{C.~L. Zitnick}, \bibinfo{author}{N.~Yakubova},
  \bibinfo{author}{F.~Knoll}, \bibinfo{author}{P.~Johnson},
\newblock \bibinfo{title}{End-to-end variational networks for accelerated {MRI}
  reconstruction},
\newblock in: \bibinfo{booktitle}{Medical Image Computing and Computer Assisted
  Intervention--MICCAI 2020: 23rd International Conference, Lima, Peru, October
  4--8, 2020, Proceedings, Part II 23}, \bibinfo{organization}{Springer},
  \bibinfo{year}{2020}, pp. \bibinfo{pages}{64--73}.
%Type = Article
\bibitem[{Fabian et~al.(2022)Fabian, Tinaz, and
  Soltanolkotabi}]{fabian2022humus}
\bibinfo{author}{Z.~Fabian}, \bibinfo{author}{B.~Tinaz},
  \bibinfo{author}{M.~Soltanolkotabi},
\newblock \bibinfo{title}{{HUMUS-Net}: Hybrid unrolled multi-scale network
  architecture for accelerated {MRI} reconstruction},
\newblock \bibinfo{journal}{Advances in Neural Information Processing Systems}
  \bibinfo{volume}{35} (\bibinfo{year}{2022}) \bibinfo{pages}{25306--25319}.
%Type = Article
\bibitem[{Vornehm et~al.(2025)Vornehm, Wetzl, Giese, F{\"u}rnrohr, Pang, Chow,
  Gebker, Ahmad, and Knoll}]{vornehm2025cinevn}
\bibinfo{author}{M.~Vornehm}, \bibinfo{author}{J.~Wetzl},
  \bibinfo{author}{D.~Giese}, \bibinfo{author}{F.~F{\"u}rnrohr},
  \bibinfo{author}{J.~Pang}, \bibinfo{author}{K.~Chow},
  \bibinfo{author}{R.~Gebker}, \bibinfo{author}{R.~Ahmad},
  \bibinfo{author}{F.~Knoll},
\newblock \bibinfo{title}{Cinevn: Variational network reconstruction for rapid
  functional cardiac cine mri},
\newblock \bibinfo{journal}{Magnetic Resonance in Medicine}
  \bibinfo{volume}{93} (\bibinfo{year}{2025}) \bibinfo{pages}{138--150}.
%Type = Article
\bibitem[{Xiang et~al.(2021)Xiang, Dong, and Yang}]{xiang2021fista}
\bibinfo{author}{J.~Xiang}, \bibinfo{author}{Y.~Dong},
  \bibinfo{author}{Y.~Yang},
\newblock \bibinfo{title}{{FISTA-Net}: Learning a fast iterative shrinkage
  thresholding network for inverse problems in imaging},
\newblock \bibinfo{journal}{IEEE Transactions on Medical Imaging}
  \bibinfo{volume}{40} (\bibinfo{year}{2021}) \bibinfo{pages}{1329--1339}.
%Type = Inproceedings
\bibitem[{Zhang et~al.(2023)Zhang, Li, and Hu}]{zhang2023tlr}
\bibinfo{author}{Y.~Zhang}, \bibinfo{author}{P.~Li}, \bibinfo{author}{Y.~Hu},
\newblock \bibinfo{title}{Dynamic {MRI} using learned transform-based tensor
  low-rank network ({LT2LR-NET})},
\newblock in: \bibinfo{booktitle}{2023 IEEE 20th International Symposium on
  Biomedical Imaging}, \bibinfo{year}{2023}, pp. \bibinfo{pages}{1--4}.
  \DOIprefix\doi{10.1109/ISBI53787.2023.10230437}.
%Type = Article
\bibitem[{Huang et~al.(2021)Huang, Ke, Cui, Cheng, Qiu, Jia, Ying, Zhu, and
  Liang}]{huang2021deep}
\bibinfo{author}{W.~Huang}, \bibinfo{author}{Z.~Ke}, \bibinfo{author}{Z.-X.
  Cui}, \bibinfo{author}{J.~Cheng}, \bibinfo{author}{Z.~Qiu},
  \bibinfo{author}{S.~Jia}, \bibinfo{author}{L.~Ying},
  \bibinfo{author}{Y.~Zhu}, \bibinfo{author}{D.~Liang},
\newblock \bibinfo{title}{Deep low-rank plus sparse network for dynamic {MR}
  imaging},
\newblock \bibinfo{journal}{Medical Image Analysis} \bibinfo{volume}{73}
  (\bibinfo{year}{2021}) \bibinfo{pages}{102190}.
%Type = Article
\bibitem[{Chen et~al.(2023)Chen, Xia, Yang, Chen, Liu, Zhou, Wang, Chen, Wen,
  and Zhang}]{chen2023soul}
\bibinfo{author}{X.~Chen}, \bibinfo{author}{W.~Xia}, \bibinfo{author}{Z.~Yang},
  \bibinfo{author}{H.~Chen}, \bibinfo{author}{Y.~Liu},
  \bibinfo{author}{J.~Zhou}, \bibinfo{author}{Z.~Wang},
  \bibinfo{author}{Y.~Chen}, \bibinfo{author}{B.~Wen},
  \bibinfo{author}{Y.~Zhang},
\newblock \bibinfo{title}{{SOUL-Net}: A sparse and low-rank unrolling network
  for spectral {CT} image reconstruction},
\newblock \bibinfo{journal}{IEEE Transactions on Neural Networks and Learning
  Systems}  (\bibinfo{year}{2023}).
%Type = Inproceedings
\bibitem[{Zhong et~al.(2024)Zhong, Huang, Li, Wang, Hu, Li, Liang, Zheng, and
  Zhang}]{zhong2024jslrnet}
\bibinfo{author}{Y.~Zhong}, \bibinfo{author}{M.~Huang},
  \bibinfo{author}{J.~Li}, \bibinfo{author}{Y.~Wang}, \bibinfo{author}{Z.~Hu},
  \bibinfo{author}{Y.~Li}, \bibinfo{author}{D.~Liang},
  \bibinfo{author}{H.~Zheng}, \bibinfo{author}{N.~Zhang},
\newblock \bibinfo{title}{{JSLRNet}: Joint sparse and low-rank unfolding
  network for {MR} image reconstruction},
\newblock in: \bibinfo{booktitle}{Medical Imaging with Deep Learning},
  \bibinfo{year}{2024}.
%Type = Article
\bibitem[{Combettes and Pesquet(2008)}]{combettes2008proximal}
\bibinfo{author}{P.~L. Combettes}, \bibinfo{author}{J.-C. Pesquet},
\newblock \bibinfo{title}{A proximal decomposition method for solving convex
  variational inverse problems},
\newblock \bibinfo{journal}{Inverse problems} \bibinfo{volume}{24}
  (\bibinfo{year}{2008}) \bibinfo{pages}{065014}.
%Type = Article
\bibitem[{Zhao et~al.(2019)Zhao, Zhong, Fu, Tang, and Pecht}]{zhao2019deep}
\bibinfo{author}{M.~Zhao}, \bibinfo{author}{S.~Zhong}, \bibinfo{author}{X.~Fu},
  \bibinfo{author}{B.~Tang}, \bibinfo{author}{M.~Pecht},
\newblock \bibinfo{title}{Deep residual shrinkage networks for fault
  diagnosis},
\newblock \bibinfo{journal}{IEEE Transactions on Industrial Informatics}
  \bibinfo{volume}{16} (\bibinfo{year}{2019}) \bibinfo{pages}{4681--4690}.
%Type = Article
\bibitem[{Kingma and Ba(2014)}]{kingma2014adam}
\bibinfo{author}{D.~P. Kingma}, \bibinfo{author}{J.~Ba},
\newblock \bibinfo{title}{{Adam}: A method for stochastic optimization},
\newblock \bibinfo{journal}{arXiv preprint arXiv:1412.6980}
  (\bibinfo{year}{2014}).
%Type = Article
\bibitem[{Zeiler(2012)}]{zeiler2012adadelta}
\bibinfo{author}{M.~D. Zeiler},
\newblock \bibinfo{title}{Adadelta: an adaptive learning rate method},
\newblock \bibinfo{journal}{arXiv preprint arXiv:1212.5701}
  (\bibinfo{year}{2012}).
%Type = Inproceedings
\bibitem[{Abadi et~al.(2016)Abadi, Barham, Chen, Chen, Davis, Dean, Devin,
  Ghemawat, Irving, Isard et~al.}]{abadi2016tensorflow}
\bibinfo{author}{M.~Abadi}, \bibinfo{author}{P.~Barham},
  \bibinfo{author}{J.~Chen}, \bibinfo{author}{Z.~Chen},
  \bibinfo{author}{A.~Davis}, \bibinfo{author}{J.~Dean},
  \bibinfo{author}{M.~Devin}, \bibinfo{author}{S.~Ghemawat},
  \bibinfo{author}{G.~Irving}, \bibinfo{author}{M.~Isard}, et~al.,
\newblock \bibinfo{title}{{TensorFlow}: A system for {Large-Scale} machine
  learning},
\newblock in: \bibinfo{booktitle}{12th USENIX symposium on operating systems
  design and implementation}, \bibinfo{year}{2016}, pp.
  \bibinfo{pages}{265--283}.
%Type = Article
\bibitem[{Chen et~al.(2020)Chen, Liu, Schniter, Tong, Zareba, Simonetti,
  Potter, and Ahmad}]{ref_ocmr}
\bibinfo{author}{C.~Chen}, \bibinfo{author}{Y.~Liu},
  \bibinfo{author}{P.~Schniter}, \bibinfo{author}{M.~Tong},
  \bibinfo{author}{K.~Zareba}, \bibinfo{author}{O.~Simonetti},
  \bibinfo{author}{L.~Potter}, \bibinfo{author}{R.~Ahmad},
\newblock \bibinfo{title}{{OCMR} (v1. 0)--open-access multi-coil k-space
  dataset for cardiovascular magnetic resonance imaging},
\newblock \bibinfo{journal}{arXiv preprint arXiv:2008.03410}
  (\bibinfo{year}{2020}).
%Type = Article
\bibitem[{Uecker et~al.(2014)Uecker, Lai, Murphy, Virtue, Elad, Pauly,
  Vasanawala, and Lustig}]{ref_multicoil}
\bibinfo{author}{M.~Uecker}, \bibinfo{author}{P.~Lai}, \bibinfo{author}{M.~J.
  Murphy}, \bibinfo{author}{P.~Virtue}, \bibinfo{author}{M.~Elad},
  \bibinfo{author}{J.~M. Pauly}, \bibinfo{author}{S.~S. Vasanawala},
  \bibinfo{author}{M.~Lustig},
\newblock \bibinfo{title}{{ESPIRiT}—an eigenvalue approach to autocalibrating
  parallel {MRI}: where {SENSE} meets {GRAPPA}},
\newblock \bibinfo{journal}{Magnetic resonance in medicine}
  \bibinfo{volume}{71} (\bibinfo{year}{2014}) \bibinfo{pages}{990--1001}.
%Type = Article
\bibitem[{Roemer et~al.(1990)Roemer, Edelstein, Hayes, Souza, and
  Mueller}]{roemer1990nmr}
\bibinfo{author}{P.~B. Roemer}, \bibinfo{author}{W.~A. Edelstein},
  \bibinfo{author}{C.~E. Hayes}, \bibinfo{author}{S.~P. Souza},
  \bibinfo{author}{O.~M. Mueller},
\newblock \bibinfo{title}{The {NMR} phased array},
\newblock \bibinfo{journal}{Magnetic resonance in medicine}
  \bibinfo{volume}{16} (\bibinfo{year}{1990}) \bibinfo{pages}{192--225}.
%Type = Article
\bibitem[{Ahmad et~al.(2015)Ahmad, Xue, Giri, Ding, Craft, and
  Simonetti}]{ahmad2015variable}
\bibinfo{author}{R.~Ahmad}, \bibinfo{author}{H.~Xue},
  \bibinfo{author}{S.~Giri}, \bibinfo{author}{Y.~Ding},
  \bibinfo{author}{J.~Craft}, \bibinfo{author}{O.~P. Simonetti},
\newblock \bibinfo{title}{Variable density incoherent spatiotemporal
  acquisition ({VISTA}) for highly accelerated cardiac {MRI}},
\newblock \bibinfo{journal}{Magnetic resonance in medicine}
  \bibinfo{volume}{74} (\bibinfo{year}{2015}) \bibinfo{pages}{1266--1278}.
%Type = Article
\bibitem[{Wang et~al.(2023)Wang, Lyu, Wang, Qin, Guo, Zhang, Yu, Li, Wang, Jin
  et~al.}]{wang2023cmrxrecon}
\bibinfo{author}{C.~Wang}, \bibinfo{author}{J.~Lyu}, \bibinfo{author}{S.~Wang},
  \bibinfo{author}{C.~Qin}, \bibinfo{author}{K.~Guo},
  \bibinfo{author}{X.~Zhang}, \bibinfo{author}{X.~Yu}, \bibinfo{author}{Y.~Li},
  \bibinfo{author}{F.~Wang}, \bibinfo{author}{J.~Jin}, et~al.,
\newblock \bibinfo{title}{{CMRxRecon}: an open cardiac {MRI} dataset for the
  competition of accelerated image reconstruction},
\newblock \bibinfo{journal}{arXiv preprint arXiv:2309.10836}
  (\bibinfo{year}{2023}).
%Type = Article
\bibitem[{Knoll et~al.(2020)Knoll, Murrell, Sriram, Yakubova, Zbontar, Rabbat,
  Defazio, Muckley, Sodickson, Zitnick et~al.}]{knoll2020advancing}
\bibinfo{author}{F.~Knoll}, \bibinfo{author}{T.~Murrell},
  \bibinfo{author}{A.~Sriram}, \bibinfo{author}{N.~Yakubova},
  \bibinfo{author}{J.~Zbontar}, \bibinfo{author}{M.~Rabbat},
  \bibinfo{author}{A.~Defazio}, \bibinfo{author}{M.~J. Muckley},
  \bibinfo{author}{D.~K. Sodickson}, \bibinfo{author}{C.~L. Zitnick}, et~al.,
\newblock \bibinfo{title}{Advancing machine learning for {MR} image
  reconstruction with an open competition: Overview of the 2019 {fastMRI}
  challenge},
\newblock \bibinfo{journal}{Magnetic resonance in medicine}
  \bibinfo{volume}{84} (\bibinfo{year}{2020}) \bibinfo{pages}{3054--3070}.
%Type = Article
\bibitem[{Zbontar et~al.(2018)Zbontar, Knoll, Sriram, Murrell, Huang, Muckley,
  Defazio, Stern, Johnson, Bruno et~al.}]{zbontar2018fastmri}
\bibinfo{author}{J.~Zbontar}, \bibinfo{author}{F.~Knoll},
  \bibinfo{author}{A.~Sriram}, \bibinfo{author}{T.~Murrell},
  \bibinfo{author}{Z.~Huang}, \bibinfo{author}{M.~J. Muckley},
  \bibinfo{author}{A.~Defazio}, \bibinfo{author}{R.~Stern},
  \bibinfo{author}{P.~Johnson}, \bibinfo{author}{M.~Bruno}, et~al.,
\newblock \bibinfo{title}{{fastMRI}: An open dataset and benchmarks for
  accelerated {MRI}},
\newblock \bibinfo{journal}{arXiv preprint arXiv:1811.08839}
  (\bibinfo{year}{2018}).
%Type = Article
\bibitem[{Wang et~al.(2021)Wang, Dang, Hu, Fua, and Salzmann}]{wang2021robust}
\bibinfo{author}{W.~Wang}, \bibinfo{author}{Z.~Dang}, \bibinfo{author}{Y.~Hu},
  \bibinfo{author}{P.~Fua}, \bibinfo{author}{M.~Salzmann},
\newblock \bibinfo{title}{Robust differentiable {SVD}},
\newblock \bibinfo{journal}{IEEE Transactions on Pattern Analysis and Machine
  Intelligence} \bibinfo{volume}{44} (\bibinfo{year}{2021})
  \bibinfo{pages}{5472--5487}.

\end{thebibliography}

\end{document}